\documentclass[a4paper,fleqn]{cas-sc}

\usepackage[authoryear,longnamesfirst]{natbib}

\ExplSyntaxOn
\cs_gset:Npn \__first_footerline: { }
\ExplSyntaxOff

\def\tsc#1{\csdef{#1}{\textsc{\lowercase{#1}}\xspace}}
\tsc{WGM}
\tsc{QE}
\tsc{EP}
\tsc{PMS}
\tsc{BEC}
\tsc{DE}

\begin{document}
\let\WriteBookmarks\relax
\def\floatpagepagefraction{1}
\def\textpagefraction{.001}
\shorttitle{}
\shortauthors{}

\title [mode = title]{Evaluating the Alignment Between GeoAI Explanations and Domain Knowledge in Satellite-Based Flood Mapping}                      

\author[1]{Hyunho Lee}[orcid=0009-0000-3172-996X]

\author[1]{Wenwen Li}[orcid=0000-0003-2237-9499]

\cormark[1]

\ead{wenwen@asu.edu}

\affiliation[1]{organization={School of Geographical Sciences and Urban Planning, Arizona State University},
            city={Tempe},
            postcode={85281}, 
            state={AZ},
            country={USA}}

\begin{abstract}
The increasing number of satellites has improved the temporal resolution of Earth observation, making satellite-based flood mapping a promising approach for operational flood monitoring. Deep learning-based approaches for flood mapping using satellite imagery, an important application within Geospatial Artificial Intelligence (GeoAI), have shown improved predictive performance by learning complex spatial and spectral patterns from large volumes of remote sensing data. However, the opaque decision-making processes of deep learning models remain a major barrier to their integration into critical scientific and operational workflows. This highlights the need for a systematic assessment of whether model explanations align with established domain knowledge in remote sensing. To address this research gap, this study introduces the ADAGE (Alignment between Domain Knowledge and GeoAI Explanation Evaluation) framework. The proposed framework is designed to systematically evaluate how well explanations of deep learning models align with established remote sensing knowledge, particularly regarding the distinctive spectral properties of the Earth's surface. The ADAGE framework employs Channel-Group SHAP (SHapley Additive exPlanations) method to estimate the contributions of grouped input channels to pixel-level predictions. Experiments on two satellite-based flood mapping tasks demonstrate that the ADAGE framework can (1) quantitatively assess the alignment between model explanations and reference explanations derived from domain knowledge, (2) reveal that different models, even when trained using the same data and model configuration, exhibit different sensitivities to the input channel groups, leading to different explanations, and (3) help domain experts identify misaligned explanations through the proposed alignment scores. This study contributes to bridging the gap between explainability and domain knowledge in GeoAI for Earth observation, enhancing the applicability of GeoAI models in scientific and operational workflows.

\end{abstract}



\maketitle

\section{Introduction}
\label{Section1}

Satellite data are an essential source for flood mapping, as they provide regular revisit intervals and enable the monitoring of vast and inaccessible areas, unlike ground-based systems such as Internet of Things (IoT) sensors \citep{sheffield2018satellite, misra2025mapping}. However, satellite observations may not always capture peak flood conditions due to limitations in temporal resolution \citep{tarpanelli2022effectiveness}. This limitation is being mitigated by the growing number of satellites, including satellite constellations \citep{sandau2010small, li2021cubesat}. As a result, satellite-based flood mapping has become a promising approach for operational flood monitoring \citep{roth2025evaluating, wagner2026fully}.

Recently, deep learning models have improved the accuracy of flood mapping by leveraging end-to-end optimization to learn complex spatial and spectral patterns from large volumes of remote sensing data \citep{dong2021monitoring, bereczky2022sentinel, li2022water, li2023u}. Specifically, deep learning-based semantic segmentation of remote sensing data (DLSS-RS), an important Geospatial Artificial Intelligence (GeoAI) approach that generates dense pixel-level predictions from multi-band imagery, is widely used to delineate flood extents \citep{bonafilia2020sen1floods11, NEURIPS2024_43612b06, zhao2024urbansarfloods, lee2026spatially}. However, despite enhanced predictive performance, the opaque nature of deep learning models’ decision-making processes remains a major barrier to their integration into critical scientific and operational workflows \citep{kotaridis2021remote, hohl2024opening}. In particular, domain experts have limited methods to assess whether the model’s decision-making process aligns with established knowledge or instead exploits spurious correlations in the training data through shortcut learning \citep{geirhos2020shortcut}. 

This challenge has led to the development of explainable GeoAI, which aims to improve understanding of GeoAI, particularly deep learning models, by estimating the contribution of input features to their predictions \citep{hsu2023explainable, li2024geoshapley}. However, current explainable GeoAI research on DLSS-RS models faces two main limitations. First, there is insufficient research on pixel-level explanations for DLSS-RS models \citep{hohl2024opening, gipivskis2024explainable}. This is due to the complexity of handling multi-dimensional inputs and dense pixel-level outputs. Second, despite the substantial body of scientific knowledge on the physical properties of Earth’s surface features in remote sensing, particularly regarding which spectral bands are relevant for detecting specific land surface conditions or environmental phenomena, there has been limited study on evaluating the alignment between GeoAI model explanations and established domain knowledge.

To address these research gaps, we propose the ADAGE (Alignment between Domain Knowledge and GeoAI Explanation Evaluation) framework, a novel GeoAI evaluation approach for DLSS-RS models. The proposed framework employs Channel-Group SHAP (SHapley Additive exPlanations) to generate pixel-level explanations whose explanatory level aligns with established domain knowledge in remote sensing through channel grouping. The alignment between these model explanations and reference explanations derived from domain knowledge is then quantitatively evaluated. We validate the proposed framework through two satellite-based flood mapping case studies using publicly available datasets: (1) a multimodal post-flood water extent mapping task using Synthetic Aperture Radar (SAR) and Multispectral Imaging (MSI) data under cloudy conditions, and (2) a task detecting open and flooded urban areas from pre-flood and post-flood SAR data. Based on the two case studies, we demonstrate that the proposed framework provides a systematic approach that enables domain experts to identify misaligned GeoAI model explanations, whether due to novel patterns or spurious correlations learned by the model.

The main contributions of this study are: 

\begin{enumerate}[(\arabic*)]

\item We propose the ADAGE (Alignment between Domain Knowledge And GeoAI Explanation Evaluation) framework to quantitatively evaluate the alignment between GeoAI explanations and reference explanations derived from domain knowledge.

\item To align the explanatory level of GeoAI with reference explanations, we introduce Channel-Group SHAP method, which extends the concept of Grouped Shapley Values (GSV) by grouping input channels to explain DLSS-RS models.

\item We demonstrate the effectiveness of the proposed framework in identifying misalignment between model explanations and reference explanations by applying it to two satellite-based flood mapping case studies.

\end{enumerate}

The structure of this paper is as follows: Section 2 reviews relevant literature; Section 3 details the proposed framework; Section 4 outlines the experimental setup, and Section 5 presents the results; Section 6 provides a discussion; Finally, Section 7 summarizes the findings and suggests directions for future research.

\section{Literature review}
\label{Section2}

\subsection{Data Sources for Satellite-Based Flood Mapping}
\label{Section2.1}

Synthetic Aperture Radar (SAR) intensity data are the primary source for satellite-based flood mapping because SAR sensors enable observations of the Earth's surface under all weather conditions, both day and night \citep{chaouch2012synergetic, boccardo2014remote, ajmar2017response, uddin2019operational}. This capability arises from the use of microwave signals, which can penetrate cloud cover and do not depend on solar illumination \citep{grimaldi2020flood}. SAR sensors receive the backscattered energy of emitted microwave pulses, whose intensity is primarily determined by surface roughness characteristics \citep{zhao2024urban}. While rough terrain surfaces generate high backscatter by dispersing energy in multiple directions including back to the sensor, open water surfaces produce low backscatter by reflecting radar signals away from the sensor. This characteristic is exploited in flood mapping, as water surfaces reflect microwave signals away from the sensor, resulting in low backscatter. However, this property also leads to difficulty in distinguishing man-made flat surfaces (e.g., tarmac) from open water, as both can exhibit similarly low backscatter responses \citep{zhao2024urbansarfloods}. In addition, in urban areas, SAR backscatter can increase significantly during floods due to strong double-bounce interactions between partially submerged building facades and the water surface  \citep{zhao2024urban}.

Interferometric SAR (InSAR) coherence is another satellite data source that is utilized in urban flood mapping \citep{li2019urban, zhao2022urban}. When buildings are flooded, the coherence between pre- and post-event radar signals decreases, providing a useful indicator for detecting floods, particularly when changes in radar intensity of the flooded buildings are too small to detect \citep{zhao2024urban}. However, this approach relies on the assumption that buildings in urban areas exhibit low variability and thus high coherence, which can be compromised by the movement of vehicles or people, as well as by new construction activities.

Multispectral imagery provides valuable water-sensitive spectral bands, such as Near Infrared (NIR) and Shortwave Infrared (SWIR), under clear-sky conditions, which significantly enhance the accuracy of flood mapping \citep{konapala2021exploring}. These bands have longer wavelengths than visible RGB bands, leading to lower scattering and higher transmittance through thin clouds, thereby enabling improved observation of the Earth's surface \citep{li2021deep, li2022thin, zhang2023thin}. In addition, they are mostly absorbed by water, resulting in low reflectance over flooded areas \citep{sun2017cloud, mondejar2019near}. However, multispectral imagery can be limited by cloud cover, particularly during the flood season when heavy rainfall is often accompanied by extensive cloudiness. Despite this limitation, multispectral imagery are predominantly used to assess flood-induced damage to infrastructure, such as buildings and roads, due to their ease of visual interpretation \citep{boccardo2014remote, ajmar2017response}.

Collectively, the characteristics of SAR intensity, InSAR coherence, and multispectral imagery indicate that satellite-based flood mapping inherently relies on data derived from multiple sensors, where each data source provides complementary information that can be jointly utilized for flood mapping. This motivates the use of deep learning models that integrate multimodal satellite data through end-to-end optimization to capture complex spatial and spectral patterns.

\subsection{Deep Learning for Satellite-Based Flood Mapping}
\label{Section2.2}

The improved predictive performance, along with the complexity and heterogeneity of satellite data sources, motivates the adoption of deep learning models for satellite-based flood mapping. To the best of our knowledge, Convolutional Neural Networks (CNNs) were first applied to satellite-based flood mapping in 2018, leveraging image pattern recognition capabilities \citep{kang2018flood, nogueira2018exploiting}. Subsequent studies have shown that CNNs outperform conventional rule-based and machine learning approaches in satellite-based flood mapping \citep{gebrehiwot2019deep, wieland2019modular, nemni2020fully, dong2021monitoring, katiyar2021near, bereczky2022sentinel, li2023u}. In addition, researchers have published benchmark datasets for satellite-based flood mapping \citep{khouakhi2022need}, highlighting the essential role of training data in the learning process of CNNs. More recently, benchmark datasets have expanded to include multimodal geospatial inputs \citep{bonafilia2020sen1floods11, rambour2020flood, Cloud_to_Street2022, drakonakis2022ombrianet, montello2022mmflood, he2023cross, zhang2023new}, time-series SAR \citep{NEURIPS2024_43612b06}, and urban flood mapping datasets \citep{zhao2024urbansarfloods}. The availability of these benchmarks have significantly accelerated GeoAI research in satellite-based flood mapping \citep{bai2021enhancement, katiyar2021near, konapala2021exploring, garg2023cross, li2023assessment, lee2024improving, wang2024multi, lee2026spatially}. 

Research on deep learning models for satellite-based flood mapping has primarily focused on two directions. The first direction emphasizes integrating the unique characteristics of geospatial data into model design. Recent models explicitly exploit multimodality \citep{kim2021synergistic, konapala2021exploring, munoz2021local, lee2026spatially}, time-series data \citep{peng2019patch, sudiana2024performance}, and varying spatial resolutions \citep{zhang2021flood}. In addition, studies proposed architectures utilizing both multimodality and time-series characteristics \citep{drakonakis2022ombrianet, he2023cross, ahmadi2026tle} to improve mapping performance. Furthermore, to alleviate the burden of labeled data annotation, weakly supervised learning \citep{bonafilia2020sen1floods11}, unsupervised learning \citep{li2019urban, akiva2021h2o}, active learning \citep{lee2024improving}, and transfer learning \citep{li2023assessment} has been explored. Moreover, extending the scope beyond CNNs, other architectures such as Vision Transformers have also been investigated for satellite-based flood mapping \citep{saleh2024dam}.

The second direction focuses on addressing practical constraints in operational flood monitoring. For example, the authors in \citep{wieland2019modular} introduced a processing chain including the essential modules for satellite-based flood mapping, demonstrating that deep learning models can be effectively integrated into operational frameworks. However, in real-world scenarios, the satellite data acquisition process often dominates the total processing time, making it necessary to anticipate image demand and request satellite observations in advance to ensure timely flood maps \citep{wania2021increasing}. To mitigate challenges related to large file sizes and limited transmission bandwidth, onboard processing approaches that transmit processed flood maps instead of raw satellite data have been proposed \citep{mateo2021towards}. These studies highlight that successful operational deployment requires careful consideration of practical constraints. 

Consequently, as deep learning models increasingly incorporate multimodal, multi-temporal, and high-dimensional geospatial inputs, their predictive performance has improved; however, their decision-making processes remain opaque, posing a major barrier to their integration into critical scientific and operational workflows. This motivates the development of evaluation frameworks that assess how well model explanations align with the domain knowledge described in Section \ref{Section2.1}.

\subsection{Evaluation of Deep Learning Model Explanations in Remote Sensing}
\label{Section2.3}

In satellite-based flood mapping using multimodal data, deep learning models exhibit strong predictive performance, but opaque decision-making processes require systematic evaluation of model explanations to understand model behavior. The authors in \citep{nauta2023anecdotal} categorized 12 model explanation evaluation criteria into three main groups. The first group, content, evaluates the intrinsic quality of the explanation, including properties such as correctness, completeness, consistency, continuity, contrastivity, and covariate complexity. These properties describe how faithfully and robustly an explanation reflects model behavior and represents relationships among input features. The second group, presentation, focuses on how explanations are conveyed, including compactness, composition, and confidence. These properties assess the size and structure of the explanation, along with the reliability of the associated confidence information. The third group, user-related properties, includes context, coherence, and controllability, and evaluates how relevant, understandable, and user-aligned the explanations are, including their alignment with prior knowledge. 

In the field of remote sensing, a number of studies have evaluated explanations of deep learning models; however, most of these studies primarily focus on evaluating intrinsic model properties such as faithfulness and robustness, which belong to the content group \citep{kakogeorgiou2021evaluating, hohl2024opening}. Specifically, in satellite-based flood mapping, studies have largely been limited to visual explanations using mechanisms such as attention or Gradient-weighted Class Activation Mapping (Grad-CAM) \citep{sanderson2023optimal, chen2024my}, without extending to the evaluation of whether these explanations align with established remote sensing knowledge as described in Section \ref{Section2.1}. Consequently, systematic assessment of the alignment between model explanations and domain knowledge remains underexplored. 

To evaluate the alignment between model explanations and domain knowledge, both should be provided at a comparable level. However, conventional feature-level explanation methods are not directly suitable for this purpose, as experts typically interpret remote sensing data through sensor modalities or band groups associated with common physical characteristics, rather than individual bands. In this context, the concept of GroupSHAP \citep{kierdorf2024investigating, jullum2021groupshapley} is particularly suitable, as it allows estimating contributions at the level of predefined feature groups, making the model explanations directly comparable with domain knowledge. GroupSHAP, an extension of the SHAP framework, provides a theoretical foundation for model interpretation at the level of feature groups by utilizing the formulation of Grouped Shapley Values (GSV) \citep{huber2023grouping}. GSV extend the original Shapley value formulation from estimating the contribution of individual features to estimating the contribution of feature groups to model predictions, while preserving the fundamental properties of the standard Shapley values. In the GroupSHAP, the unit of model explanation is a feature group rather than an individual feature, and feature groups must be predefined as mutually exclusive subsets that collectively cover the entire feature set. As a result, the GroupSHAP is computationally more efficient than standard SHAP, since it considers combinations of feature groups rather than individual features. However, this concept has not yet been applied for interpreting DLSS-RS models, and studies evaluating alignment between GeoAI model explanations and domain knowledge remain limited.

Overall, the literature indicates that satellite-based flood mapping relies on heterogeneous geospatial data sources, each with distinct physical sensitivities and limitations, while deep learning models have increasingly leveraged these complex inputs to improve predictive performance, particularly through multimodal approaches. Despite this progress, understanding of the models’ decision-making processes remains limited, and the alignment between model explanations and established remote sensing knowledge has not been sufficiently examined. This suggests the importance of evaluation frameworks capable of assessing model explanations at a domain-relevant level of abstraction, such as sensor or band groups, rather than relying solely on generic saliency maps.

\section{Methods}
\label{Section3}

\subsection{ADAGE Framework}
\label{Section3.1}

\begin{figure}[]
  \centering
  \includegraphics[width=1.0 \linewidth]{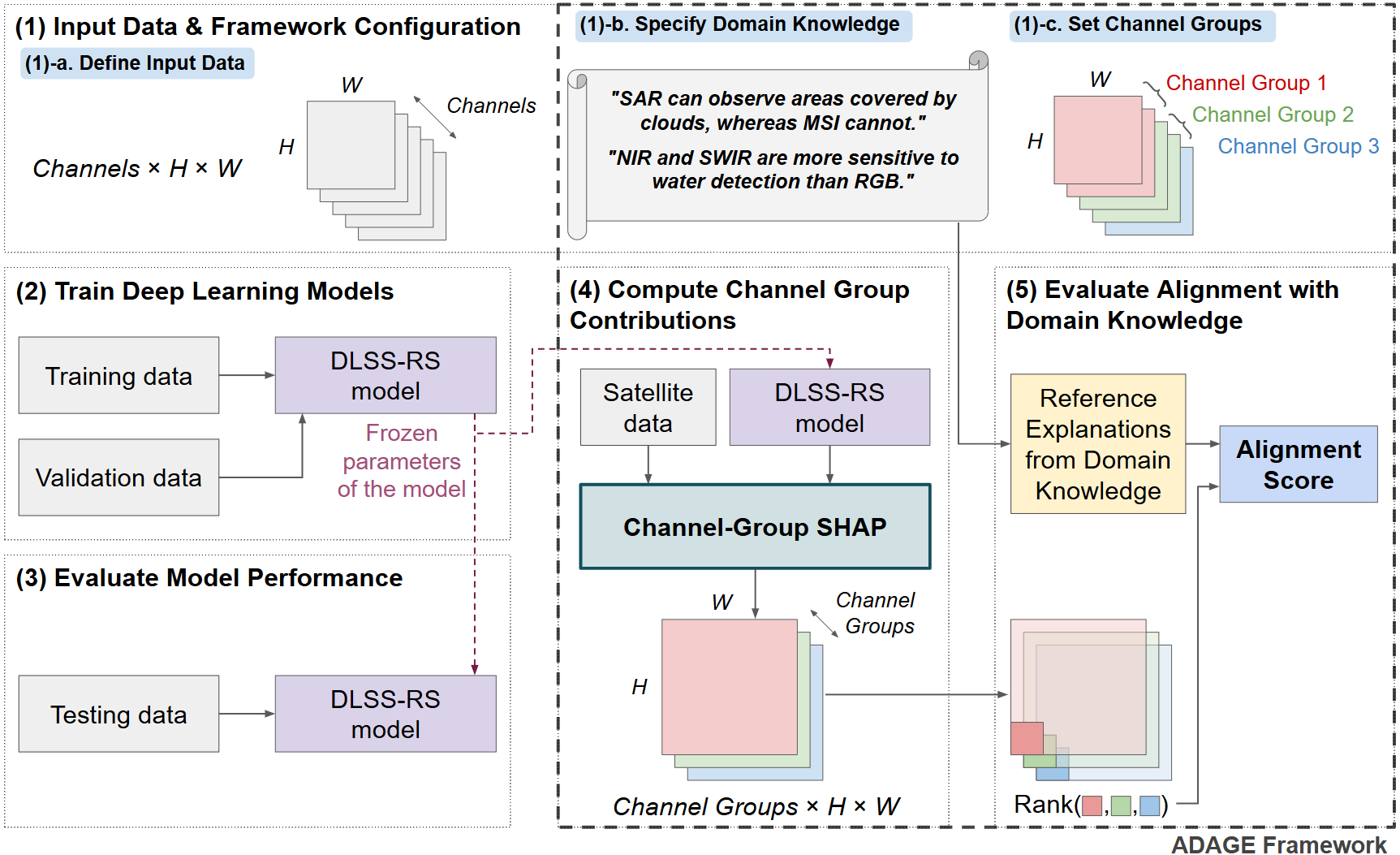}
  \caption{Overview of the ADAGE (Alignment between Domain Knowledge And GeoAI Explanation Evaluation) framework.}
  \label{fig1}
\end{figure}

In this study, we introduce the ADAGE (Alignment between Domain Knowledge And GeoAI Explanation Evaluation) framework, which is designed to systematically evaluate the alignment between GeoAI model explanations and reference explanations derived from domain knowledge in DLSS-RS. The proposed framework consists of five stages, as shown in Fig. \ref{fig1}. The first stage defines the input data of the DLSS-RS model and the configuration of the ADAGE framework. Specifically, this stage defines the number of channels, height, and width of the input data, and specifies the domain knowledge that explains the mapping outcomes. The specified domain knowledge is utilized as the reference explanation in the evaluation stage. Subsequently, channel groups are defined according to the explanatory level used in the previously specified domain knowledge. These steps enable the calculation of the alignment score between the model explanations and the reference explanation in the evaluation stage by aligning their explanatory levels.

For instance, in flood mapping, suppose the input consists of six bands from two sensors, which are SAR (VV and VH bands) and MSI (red, green, blue, and NIR bands). Regarding the characteristics of these two sensors, it is well-established domain knowledge that SAR sensors can observe the Earth's surface even in the presence of clouds \citep{zhao2024urban}, whereas MSI sensors cannot detect areas obscured by clouds. In addition, under thin cloud conditions, the NIR band is more effective for flood detection than visible light bands due to its longer wavelength \citep{li2021deep, li2022thin, zhang2023thin}. In the ADAGE framework, when this domain knowledge is used as the reference explanation, the channels are grouped into a SAR channel group (VV and VH bands), a visible light channel group (red, green, and blue bands), and a NIR channel group, reflecting the explanatory units in the domain knowledge. 

The second and third stages correspond to the training and performance evaluation of a DLSS-RS model using supervised learning. In the fourth stage, Channel-Group SHAP method is leveraged to estimate the contributions of each channel group to the predicted pixels. This Channel-Group SHAP is a method that provides explanations for channel groups in the DLSS-RS models by applying the GSV formulation within the SHAP framework. The formulation and approximation of Channel-Group SHAP are described in detail in Section \ref{Section3.2}. In the final stage, the alignment score between the reference explanation derived from domain knowledge and the model explanation estimated using Channel-Group SHAP is calculated. This alignment score enables the evaluation of the alignment between the model explanations and the domain knowledge. Further details regarding the alignment score are provided in Section \ref{Section3.3}.

\subsection{Channel-Group SHAP}
\label{Section3.2}

\subsubsection{Definition of DLSS-RS Model}
\label{Section3.2.1}

Let $f(x)$ be a deep learning-based semantic segmentation model that maps input data $x \in \mathbb{R}^{C \times H \times W}$ to an output logit map $y \in \mathbb{R}^{n_{class} \times H \times W}$, where $C$ denotes the number of input channels and $n_{class}$ denotes the number of output classes. In the output logit map, the logit corresponding to class $cls$ at spatial position (h, w) is denoted by $f_{cls,h,w}(x)$.

\subsubsection{Definition of Channel Group}
\label{Section3.2.2}
To compute the contributions of channel groups to each logit in the DLSS-RS model output, the channel groups must be defined properly. Let the complete set of channel indices be denoted by $C_{all} = \{1,2, \ldots, C\}$. A valid definition of the channel group set $F = \{cg_1, cg_2, \ldots, cg_K\}$, forming a partition of Call, must satisfy the following two conditions:

\begin{enumerate}
  \item Mutual exclusivity: $cg_i \cap cg_j = \emptyset$ for all $i \neq j$.
  \item Completeness: $\bigcup_{i=1}^{K} cg_i = C_{all}$.
\end{enumerate}

This definition extends the principles of Group Shapley Value \citep{jullum2021groupshapley, huber2023grouping}, which originally focused on feature groups, to the concept of channel groups in semantic segmentation models for environmental mapping. Channel-Group SHAP is formulated based on this extended concept within the SHAP framework. By adjusting the granularity of explanations through channel grouping, direct evaluation between reference explanations obtained from domain knowledge and explanations derived from DLSS-RS model is enabled.

\subsubsection{Channel-Group SHAP Formulation}
\label{Section3.2.3}

Based on the SHAP (SHapley Addictive exPlanations) framework, the mapping function $h_x$ transforms a binary vector of input channel groups $z' \in \{0, 1\}^{\lvert F \rvert}$ into the model’s original input space $z_S$ (see Eq. (\ref{eq1})), where $z_S$ represents an input that has missing values for channel groups not included in the subset $S$ \citep{lundberg2017unified}. $\lvert F \rvert$ denotes the total number of channel groups contained in the set $F$.

\begin{equation}
\label{eq1}
    h_x(z') = z_S.
\end{equation}

For a specific channel group $cg \in F$, its Channel-Group SHAP value $\phi_{cg,cls,h,w}$ is defined as:

\begin{equation}
\label{eq2}
\phi_{cg,cls,h,w} =
\sum_{S \subseteq F \setminus \{cg\}} w(S)
\Big[
f_{cls,h,w}(z_{S \cup \{cg\}}) - f_{cls,h,w}(z_S)
\Big],
\end{equation}

\begin{equation}
w(S) = \frac{|S|! (|F| - |S| - 1)!}{|F|!},
\end{equation}

where $F$ is the set of all channel groups, and $S$ is a subset of channel groups that excludes the channel group $cg$. $\lvert S \rvert $ denotes the size of $S$. $f_{cls,h,w}(z_{S \cup \{cg\}})$ represents the model output when the channel group $cg$ is added to the subset $S$, whereas $f_{cls,h,w}(z_S)$ corresponds to the model output when the input contains only the channel groups in $S$, excluding $cg$. Notably, the computational cost of Channel-Group SHAP increases exponentially with the number of channel groups; however, it can be managed through appropriate channel grouping strategies \citep{jullum2021groupshapley, huber2023grouping}.

\subsubsection{Channel-Group SHAP Approximation}
\label{Section3.2.4}

In Channel-Group SHAP formula Eq. (\ref{eq2}), the primary computational challenge arises from evaluating $f_{cls,h,w}(z_S)$, as most deep learning-based semantic segmentation models are not designed to handle arbitrary patterns of missing input values. To address this issue, we approximate $f_{cls,h,w}(z_S)$ with the conditional expectation $\mathbb{E}[f_{cls,h,w}(z) \mid z_S]$. To make this conditional expectation tractable, we further approximate the expectation under two key assumptions. First, we assume that each channel group is independent of the others. Under this assumption, the conditional expectation can be approximated as shown in Eq. (\ref{eq3}), (\ref{eq4}), and (\ref{eq5}) \citep{lundberg2017unified}.

\begin{align}
    f_{cls,h,w}(z_S)        &\approx \mathbb{E}[f_{cls,h,w}(z) \mid z_S]  \label{eq3} \\
                            &= \mathbb{E}_{z_{\tilde{S}} \mid z_S}[f_{cls,h,w}(z)] \label{eq4} \\
                            &\approx \mathbb{E}_{z_{\tilde{S}}}[f_{cls,h,w}(z)]. \label{eq5}
\end{align}

Second, we assume local linearity of the model $f_{cls,h,w}$, which posits that the model behaves approximately linearly around a reference point $x_0$. Based on this assumption, we apply a first order Taylor expansion at the channel group level, allowing us to express $f_{cls,h,w}(x)$ as follows \citep{simonyan2013deep}:

\begin{equation}
\label{eq6}
     f_{cls,h,w}(x) \approx f_{cls,h,w}(x_0) + \sum_{cg \in F} \nabla_{x_{cg}}f_{cls,h,w}(x_0)(x_{cg} - x_{0,cg}),
\end{equation}
where $x_{cg}$ denotes the vectorized form of the channel group $cg$ in the input $x$, $\nabla_{x_{cg}}f_{cls,h,w}(x_0)$ is the gradient of $f_{cls,h,w}$ with respect to $x_{cg}$ at $x_0$, $x_{0, cg}$ is the vectorized form of the channel group $cg$ in the reference point $x_0$. By combining these two assumptions, we can approximate the expectation over the absent channel group $\tilde{S}$ 

\begin{align}
    \mathbb{E}_{z_{\tilde{S}}}[f_{cls,h,w}(z)] &\approx f_{cls,h,w}(x_0) + \sum_{cg \in S} \nabla_{z_{S,cg}}f_{cls,h,w}(x_0)(z_{S,cg} - x_{0,cg}) \nonumber \\ 
    &+ \sum_{cg \in \tilde{S}} \nabla_{z_{\tilde{S},cg}}f_{cls,h,w}(x_0)(\mathbb{E}[z_{\tilde{S},cg}] - x_{0,cg}) \label{eq7} \\
    &\approx f_{cls,h,w}([z_S, \mathbb{E}[z_{\tilde{S}}]]),  \label{eq8}
\end{align}
where $z_{S,cg}$ is the vectorized form of the channel group $cg$ in the $z_S$, $\mathbb{E}[z_{\tilde{S}, cg}]$ is the expected value of the channel group $cg$ in the$z_{\tilde{S}}$. Therefore, the approximation of $f_{cls,h,w}(z_S)$ can be expressed as $f_{cls,h,w}([z_S, \mathbb{E}[z_{\tilde{S}}]])$ for use in Channel-Group SHAP value formula \citep{lundberg2017unified}:

\begin{align}
\label{eq9}
f_{cls,h,w}(z_S) \approx f_{cls,h,w}([z_S, \mathbb{E}[z_{\tilde{S}}]]).
\end{align}

Note that the reference point $x_0$ is only used to enable the first-order Taylor approximation and does not appear in the final computation of Channel-Group SHAP values. The actual attribution is derived from direct model evaluation with input where the missing channel groups are replaced by their expected values, i.e., $[z_S, \mathbb{E}[z_{\tilde{S}}]]$. Furthermore, since the estimated Channel-Group value is an approximation of the channel groups’ contribution to the logit, it should be interpreted with caution, considering the underlying assumptions and potential approximation errors.

\subsection{Alignment Score}
\label{Section3.3}

In previous studies, Intersection over Union or (rank) correlation has primarily been utilized to measure the alignment between model and reference explanations \citep{nauta2023anecdotal}. However, in the ADAGE framework, it was necessary to consider the characteristics of model explanations and reference explanations when estimating their alignment. Specifically, Channel-Group SHAP method quantifies the contribution of each channel group to the prediction at each pixel. Considering the approximation steps in Channel-Group SHAP method, using the rank of contribution for each channel group can be more robust than relying on the contribution values themselves when interpreting the importance of channel groups. In contrast, assigning precise importance rankings to reference explanations is inherently difficult because domain knowledge is typically descriptive rather than quantitative in expressing the importance of spectral features for detecting target classes (e.g., forests, water, buildings). Therefore, taking into account the descriptive nature of domain knowledge, it is more appropriate to represent the reference explanation as a set without ranking.

When the reference explanation is represented as a set of channel groups without ranking, and the model explanation provides a ranked list of channel groups, evaluating their alignment becomes analogous to a relevance matching problem in information retrieval. Therefore, in this study, we employ mean Average Precision at k (mAP@k) to quantitatively measure the alignment between the reference explanations and model explanations. Specifically, mAP@k measures the alignment of a ranked list of channel groups, provided as model explanations based on their contribution, against a set of reference explanations derived from domain knowledge.

For a single instance (e.g., a single pixel's explanation), the precision at rank $i$, denoted as $P(i)$, is defined as:
\begin{equation}
\label{eq10}
     P(i) = \frac{1}{i}\sum_{j=1}^i 1\{r_j \in G\},
\end{equation}
where $r_j$ is the channel group ranked at position $j$, $G$ is the set of channel groups used as reference explanations, $1\{r_j \in G\}$ is an indicator function that returns 1 if $r_j$ is in the reference explanations set G, and 0 otherwise. Using precision at each relevant position, Average Precision at k (AP@k) is defined as the average of the precisions at ranks where relevant times appear, up to rank k:
\begin{equation}
\label{eq11}
     \text{AP@k} = \frac{1}{\text{min}(\lvert G \rvert, k)}\sum_{i=1}^k P(i) \cdot 1\{r_j \in G\}.
\end{equation}

AP@k represents how well the top-k predictions align with the reference explanations, considering both precision and ranking order. Finally, the mean Average Precision at k (mAP@k) is the mean of AP@k over all instances (e.g., all pixels):

\begin{equation}
\label{eq12}
     \text{mAP@k} = \frac{1}{N}\sum_{n=1}^N \text{AP@k}^{(n)},
\end{equation}
where $N$ is the total number of evaluated instances, and $\text{AP@k}^{(n)}$ is the AP@k for the n-th instance. This metric provides a quantitative evaluation of how well the GeoAI model explanations align with reference explanations derived from domain knowledge. In addition, mAP@k is a robust metric for evaluating the alignment between model explanations and reference explanations because mAP@k utilizes the ranking of channel group contributions rather than their exact values, thereby reducing sensitivity to approximation errors from Channel-Group SHAP. Moreover, by averaging alignment scores across all pixels, mAP@k provides a statistically aggregated measure, robust to local variations and noise.

\section{Experimental Setup}
\label{Section4}

\subsection{Datasets}
\label{Section4.1}

This study utilized two satellite-based flood mapping datasets to demonstrate the effectiveness of the ADAGE framework in evaluating the alignment between model explanations and reference explanations derived from domain knowledge. For the first case study, we employed the C2S-MS (Cloud to Street–Microsoft) Floods dataset \citep{Cloud_to_Street2022} to demonstrate the applicability of the ADAGE framework in a multimodal post-flood water extent mapping case study. The dataset includes 900 paired Synthetic Aperture Radar (SAR) and Multispectral Imaging (MSI) images (512 $\times$ 512 pixels) from 18 global flood events between 2016 and 2020. SAR data from Sentinel-1 (VV and VH bands) and MSI data from Sentinel-2 (13 bands) were acquired over the same locations within four days after flood events. Both SAR and MSI bands were resampled to 10m resolution. For input bands, VV and VH bands from SAR and four bands (Red, Green, Blue, NIR) from MSI were used. The dataset was split into training, validation, and test sets (60:20:20) through stratified random sampling. Each 512 × 512 image in the validation and test sets was divided into four non-overlapping 256 $\times$ 256 patches, resulting in 720 samples in both validation and test datasets. Band-wise normalization was applied using the mean and standard deviation derived from the training dataset. 

For the second case study, the UrbanSARFloods dataset \citep{zhao2024urbansarfloods} was utilized. This dataset contains 8,879 paired SAR intensity and interferometric coherence data (512 $\times$ 512 pixels) from 18 global flood events between 2016 and 2023 acquired pre-flood and post-flood events. Both the SAR intensity and interferometric coherence data have a spatial resolution of 20 meters. The dataset was divided into training, validation, and test sets following the split provided by the UrbanSARFloods dataset. The training data were labeled using rule-based methods grounded in domain knowledge, whereas the test data were manually annotated by experts to provide accurate reference labels. The labels are categorized into three classes: NF (Not Flooded), FO (Flooded Open), and FU (Flooded Urban). For training data, SAR intensity was used to annotate the FO class, while interferometric coherence and the WSF (World Settlement Footprint) 2019 data were used to annotate the FU class \citep{zhao2024urbansarfloods}. However, since the UrbanSARFloods dataset contains only SAR intensity and interferometric coherence data, the WSF 2019 data corresponding to each patch were added as an input in this case study to include all data used in the label annotation process. Each 512 $\times$ 512 image in the validation and test sets was divided into four non-overlapping 256 $\times$ 256 patches, and band-wise normalization was applied using the mean and standard deviation derived from the training dataset.

\subsection{ADAGE Framework Configurations}
\label{Section4.2}

This section describes the configuration of the ADAGE framework for two satellite-based flood mapping case studies. A key step in this framework is the proper definition of channel groups, as evaluating the alignment between model explanations and reference explanations requires that both share the same level of detail. In this context, the configuration of channel groups depends on two factors: the channels available in the training dataset and the domain knowledge established for the specific environmental mapping task. The configurations for the two case studies are presented in Tables \ref{tab:tab3} and \ref{tab:tab4}.

Notably, alignment with domain knowledge evaluates the coherence of explanations with respect to reference explanations \citep{nauta2023anecdotal}. This coherence is most relevant for True Positives (TPs), where the model makes correct predictions and the alignment score can be quantitatively evaluated. In these cases, a high alignment score indicates that the model is not only correct but also relies on reasoning consistent with domain knowledge. For False Positives (FPs) and False Negatives (FNs), where the prediction is incorrect, it is still possible to estimate the contributions of channel groups to the prediction. However, for incorrect predictions, quantitatively evaluating the alignment between the generated explanations and domain knowledge is more challenging because defining an appropriate quantitative evaluation criterion is not straightforward.
Therefore, this study focuses on evaluating alignment scores for TPs to examine whether the model makes correct predictions grounded in explanations consistent with domain knowledge.

In Table \ref{tab:tab3}, the input bands are grouped based on their cloud-penetrating capability: $\text{CG}_\text{SAR}$ includes both VV and VH polarizations; $\text{CG}_\text{RGB}$ includes the visible bands; and $\text{CG}_\text{NIR}$ includes the near-infrared band. These channel groups are also used as elements of the reference explanations. For pixels correctly identified as water within a cloud-labeled area, the reference explanation is defined as \{$\text{CG}_\text{SAR}$\}. This is based on the domain knowledge $\text{DKS}_\text{case1-1}$ that the SAR sensor can penetrate clouds. In addition, if the Near-Infrared (NIR) reflectance of that pixel is also below a specified threshold, the reference explanation is expanded to \{$\text{CG}_\text{SAR}$, $\text{CG}_\text{NIR}$\}, incorporating the additional domain knowledge $\text{DKS}_\text{case1-2}$ that NIR is effective for detecting water even under thin cloud conditions.

In Table \ref{tab:tab4}, the input bands are grouped according to data characteristics. $\text{CG}_\text{(Int, VV)}$ and $\text{CG}_\text{(Int, VH)}$ contain the pre- and post-event intensity for VV and VH polarizations, respectively. $\text{CG}_\text{(Coh, VV)}$ and $\text{CG}_\text{(Coh, VH)}$ include the coherence data, and $\text{CG}_\text{WSF}$ represents the channel group of the World Settlement Footprint 2019 data. These 5 channel groups are utilized in the reference explanations. Specifically, for a pixel correctly identified as an flooded open area, the reference explanation is defined as \{$\text{CG}_\text{(Int, VV)}$, $\text{CG}_\text{(Int, VH)}$\}. This is based on the domain knowledge $\text{DKS}_\text{case2-1}$ that flooded open areas are extracted using SAR intensity imagery. Furthermore, for a pixel correctly identified as an flooded urban area, the reference explanation is set to \{$\text{CG}_\text{(Coh, VV)}$, $\text{CG}_\text{(Coh, VH)}$, $\text{CG}_\text{WSF}$\}. This result incorporates the additional domain knowledge $\text{DKS}_\text{case2-2}$, which detects urban floods using changes in coherence and distinguishes built-up from non-built-up areas using the WSF data.

\clearpage
\begin{table}[t]
\caption{Configuration of the ADAGE framework for the multimodal post-flood water extent mapping case study.}
\label{tab:tab3}
\resizebox{0.8\columnwidth}{!}{%
\begin{tabular}{@{}ll@{}}
\toprule
\textbf{\begin{tabular}[c]{@{}l@{}}Configuration\\ Parameter\end{tabular}}        & \textbf{Specification}                                                                                                                                                                                                                                                                                                                                                                                                                                                                                                                                                                                                                                                                                                                                                                                                                                                                                                                                                                                        \\ \midrule \midrule
Dataset                                                                           & C2S-MS Floods dataset                                                                                                                                                                                                                                                                                                                                                                                                                                                                                                                                                                                                                                                                                                                                                                                                                                                                                                                                                                                         \\ \midrule
Input bands                                                                       & 6 bands of SAR (VV, VH) and MSI (Red, Green, Blue, NIR)                                                                                                                                                                                                                                                                                                                                                                                                                                                                                                                                                                                                                                                                                                                                                                                                                                                                                                                                                       \\ \midrule
\begin{tabular}[c]{@{}l@{}}Domain \\ Knowledge \\ Statement \\ (DKS)\end{tabular} & \begin{tabular}[c]{@{}l@{}}Regarding multimodal post-flood water extent mapping in cloud-covered areas, \vspace{0.2cm} \\ (1) $\text{DKS}_{\text{case1-1}}$: Synthetic Aperture Radar (SAR), as an active microwave \\ sensor, can observe the Earth's surface covered by clouds, whereas the \\ Multispectral Instrument (MSI), a passive optical sensor, is obstructed by clouds \\ \citep{ajmar2017response, uddin2019operational, grimaldi2020flood}. \vspace{0.2cm} \\ (2) $\text{DKS}_{\text{case1-2}}$: Under thin cloud conditions, NIR is more effective for flood detection \\ than RGB due to its longer wavelength \citep{li2021deep, zhang2023thin}.\end{tabular}                                                                                                                                                                                                                                                                                                                                                                                                                                                                                                                      \\ \midrule
\begin{tabular}[c]{@{}l@{}}Channel \\ groups\end{tabular}                         & (1) $\text{CG}_{\text{SAR}}$ = \{VV, VH\}, (2) $\text{CG}_{\text{RGB}}$ = \{Red, Green, Blue\}, (3) $\text{CG}_{\text{NIR}}$ = \{NIR\}                                                                                                                                                                                                                                                                                                                                                                                                                                                                                                                                                                                                                                                                                                                                                                                                                                                                                                                                 \\ \midrule
\begin{tabular}[c]{@{}l@{}}Reference \\ Explanations\\ (RE)\end{tabular}          & \begin{tabular}[c]{@{}l@{}}Let $p$ denote a pixel at a spatial location ($i$, $j$) shared across all input bands. \vspace{0.2cm} \\ (1) From $\text{DKS}_{\text{case1-1}}$: If a pixel $p$ is located within both the cloud-labeled region \\ and the true positive water area, the reference explanation is \{$\text{CG}_{\text{SAR}}$\}. \\ $ \quad \mapsto \quad (\ensuremath{p} \in L_{cloud}) \wedge
 (p \in \text{TP}_{\text{water}}) \Rightarrow \text{RE}_{\text{case1-1}} = \{\text{CG}_{\text{SAR}}\}$ \vspace{0.2cm} \\ (2) From $\text{DKS}_{\text{case1-1}}$ and $\text{DKS}_{\text{case1-2}}$: If a pixel $p$ is located in a cloud-labeled \\ area, correctly identified as water, and its NIR reflectance is below the defined \\ threshold, then the reference explanation is \{$\text{CG}_{\text{SAR}}$, $\text{CG}_{\text{NIR}}$\}.\\ $ \quad \mapsto \quad (\ensuremath{p} \in L_{cloud}) \wedge (p \in \text{TP}_{\text{water}}) \wedge (R_p(\text{NIR}) < T_{\text{NIR}}) \Rightarrow \text{RE}_{\text{case1-2}} = \{\text{CG}_{\text{SAR}}, \text{CG}_{\text{NIR}}\}$ \vspace{0.2cm} \\ \hspace{0.2cm} \raisebox{0.25ex}{\tiny$\bullet$} $L_{cloud}$: set of pixels labeled as cloud based on a human-annotated mask derived \\ \hspace{0.2cm} from the multispectral image \\  \hspace{0.2cm} \raisebox{0.25ex}{\tiny$\bullet$} $\text{TP}_\text{water}$: set of pixels correctly classified as water \\  \hspace{0.2cm} \raisebox{0.25ex}{\tiny$\bullet$} $R_p (\text{CG}_X)$: reflectance of $\text{CG}_X$ at pixel $p$ \\ \hspace{0.2cm} \raisebox{0.25ex}{\tiny$\bullet$} $T_{\text{NIR}}$: threshold value for NIR reflectance (set to 0.2 in this study)\end{tabular} \\ \bottomrule
\end{tabular}%
}
\end{table}
\clearpage


\begin{table}[t]
\caption{Configuration of the ADAGE framework for the SAR-based open and urban area flood mapping case study.}
\label{tab:tab4}
\resizebox{0.8\columnwidth}{!}{%
\begin{tabular}{@{}ll@{}}
\toprule
\textbf{\begin{tabular}[c]{@{}l@{}}Configuration\\ Parameter\end{tabular}}        & \textbf{Specification}                                                                                                                                                                                                                                                                                                                                                                                                                                                                                                                                                                                                                                                                                                                                    \\ \midrule \midrule
Dataset                                                                           & UrbanSARFloods dataset augmented with WSF 2019 data                                                                                                                                                                                                                                                                                                                                                                                                                                                                                                                                                                                                                                                                                                       \\ \midrule
Input bands                                                                       & \begin{tabular}[c]{@{}l@{}}9 bands of VV coherence (pre- and post- event), VH coherence (pre- \\ and post- event), VV intensity (pre- and post- event), VH intensity \\ (pre- and post- event), and WSF 2019\end{tabular}                                                                                                                                                                                                                                                                                                                                                                                                                                                                                                                                 \\ \midrule
\begin{tabular}[c]{@{}l@{}}Domain \\ Knowledge \\ Statement \\ (DKS)\end{tabular} & \begin{tabular}[c]{@{}l@{}}The semi-automatic labeling rules for the UrbanSARFloods dataset \\ are as follows \citep{zhao2024urbansarfloods}. \vspace{0.2cm} \\ (1) $\text{DKS}_{\text{case2-1}}$: The flooded open areas are extracted by applying a \\ hierarchical Split-based change detection approach (i.e., HSBA) to \\ SAR intensity imagery. \vspace{0.2cm} \\ (2) $\text{DKS}_{\text{case2-2}}$: Urban floods are extracted by applying a threshold \\ (fixed at 0.3 based on trial and error) to the difference in interferometric \\ coherence images (i.e., pre-event coherence minus co-event coherence). \\ Built-up and non-built-up areas are distinguished using the World \\ Settlement Footprint 2019 (WSF2019). \vspace{0.2cm} \\ (3) $\text{DKS}_{\text{case2-3}}$: If a pixel is identified as flooded in either the VV or VH \\ polarization, it is labeled as flooded.\end{tabular} \\ \midrule
\begin{tabular}[c]{@{}l@{}}Channel \\ groups\end{tabular}                         & \begin{tabular}[c]{@{}l@{}}(1) $\text{CG}_{\text{(Int, VV)}}    = \{\text{Intensity}_{\text{VV}}^{\text{Pre}}, \text{Intensity}_{\text{VV}}^{\text{Post}}\}$, \\ (2) $\text{CG}_{\text{(Int, VH)}}    = \{\text{Intensity}_{\text{VH}}^{\text{Pre}}, \text{Intensity}_{\text{VH}}^{\text{Post}}\}$, \\ (3) $\text{CG}_{\text{(Coh, VV)}}    = \{\text{Coherence}_{\text{VV}}^{\text{Pre}}, \text{Coherence}_{\text{VV}}^{\text{Post}}\}$, \\ (4) $\text{CG}_{\text{(Coh, VH)}}    = \{\text{Coherence}_{\text{VH}}^{\text{Pre}}, \text{Coherence}_{\text{VH}}^{\text{Post}}\}$, \\ (5) $\text{CG}_{\text{WSF}}    = \{ \text{WSF2019} \}$\end{tabular}                                                                                                                                  \\ \midrule
\begin{tabular}[c]{@{}l@{}}Reference \\ Explanations\\ (RE)\end{tabular}          & \begin{tabular}[c]{@{}l@{}} Let $p$ denote a pixel at a spatial location ($i$, $j$) shared across all input bands. \vspace{0.2cm} \\ From $\text{DKS}_{\text{case2-1}}$ and $\text{DKS}_{\text{case2-3}}$: If a pixel $p$ is located in the true positive \\ flooded open area, the reference explanation is $\{\text{CG}_{\text{(Int, VV)}}, \text{CG}_{\text{(Int, VH)}}\}$.\\ $ \quad \mapsto \quad p \in \text{TP}_{\text{flooded\_open}} \Rightarrow \text{RE}_{\text{case2-1}} = \{\text{CG}_{\text{(Int, VV)}}, \text{CG}_{\text{(Int, VH)}}\}$ \vspace{0.2cm} \\ From $\text{DKS}_{\text{case2-2}}$ and $\text{DKS}_{\text{case2-3}}$:  If a pixel $p$ is located in the true positive \\ flooded urban area, the reference explanation is $\{\text{CG}_{\text{(Coh, VV)}}, \text{CG}_{\text{(Coh, VH)}},$ \\ 
$\text{CG}_{\text{WSF}}\}$.\\ $ \quad \mapsto \quad p \in \text{TP}_{\text{flooded\_urban}}) \Rightarrow \text{RE}_{\text{case2-2}} = \{ \text{CG}_{\text{(Coh, VV)}}, \text{CG}_{\text{(Coh, VH)}}, \text{CG}_{\text{WSF}} \}$ \vspace{0.2cm} \\ \hspace{1em} \raisebox{0.25ex}{\tiny$\bullet$} $\text{TP}_\text{flooded\_open}$: set of pixels correctly identified as flooded open area \\  \hspace{1em} \raisebox{0.25ex}{\tiny$\bullet$} $\text{TP}_\text{flooded\_urban}$: set of pixels correctly identified as flooded urban area \end{tabular} \\ \bottomrule
\end{tabular}%
}
\end{table}
\clearpage

\subsection{Implementation Details}
\label{Section4.3}

For both Case Study 1 and Case Study 2, U-Net \citep{ronneberger2015u} and U-Net++ \citep{zhou2019unetplusplus} with a ResNet-50 backbone encoder, as well as SegFormer \citep{xie2021segformer} with a Mix Transformer-B3 (MiT-B3) encoder, were implemented using the PyTorch framework. The U-Net, U-Net++, and SegFormer models used in this study have 32.5M, 49.0M, and 44.6M trainable parameters, respectively. In addition, for Case Study 1, which focuses on SAR and multispectral imagery fusion, two recent state-of-the-art models, PyramidMamba \citep{wang2025pyramidmamba} (125M trainable parameters) and SMAGNet \citep{lee2026spatially} (56M trainable parameters), were included to validate the applicability of the proposed framework to recent remote sensing segmentation architectures. 

All experiments were conducted on a workstation equipped with an NVIDIA RTX A5000 GPU and 251 GB of memory. All models were trained using the Adam optimizer \citep{kingma2014adam} with a weight decay of 0.0 and a batch size of 16. An initial learning rate of $5 \times 10^{-4}$ was used for all models except PyramidMamba, for which the learning rate was set to $5 \times 10^{-5}$ to ensure stable training, based on the training settings reported in \citep{wang2025pyramidmamba}. The number of training epochs was determined based on the convergence trends observed for each dataset. The multimodal post-flood water extent mapping case study used 200 epochs, and the SAR-based open and urban area flood mapping case study used 100 epochs. Data augmentation techniques, including random cropping and random flipping, were applied in all experiments. Binary cross entropy was employed as the loss function. All model weights were trained from scratch. The final model was selected as the one that achieved the lowest validation loss during training.

\subsection{Evaluation Metrics}
\label{Section4.4}

This study employs four metrics for evaluating predictive performance and one metric for assessing explanation alignment. The four metrics for evaluating predictive performance are Intersection over Union (IoU), Precision, Recall, and F1-score. These metrics are calculated based on the True Positive (TP), False Positive (FP), False Negative (FN), and True Negative (TN) from the confusion matrix. The definitions of each performance metric are provided in Eq. (\ref{eq13}) to Eq. (\ref{eq16}). 

\begin{equation}
\label{eq13}
     \text{IoU} = \frac{\text{TP}}{\text{TP} + \text{FP} + \text{FN}}.
\end{equation}

\begin{equation}
\label{eq14}
     \text{Precision} = \frac{\text{TP}}{\text{TP} + \text{FP}}.
\end{equation}

\begin{equation}
\label{eq15}
     \text{Recall} = \frac{\text{TP}}{\text{TP} + \text{FN}}.
\end{equation}

\begin{equation}
\label{eq16}
     \text{F1-score} = 2 \times \frac{\text{Precision} \times \text{Recall}}{\text{Precision} + \text{Recall}}.
\end{equation}

In addition, for True Positives (TPs), explanation alignment is evaluated using mAP@k, as described in Section \ref{Section3.3}. This metric measures how well the top k channel groups, identified by Channel-Group SHAP as most contributing to each pixel prediction, correspond to the channel groups in the reference explanations. Here, k is equal to the total number of channel groups in the reference explanation. Thus, the alignment score provides a quantitative measure of the consistency between model explanations and reference explanations, which are derived from established scientific knowledge, for correct predictions.

\section{Results}
\label{Section5}

\subsection{Case Study 1: Multimodal Post-flood Water Extent Mapping}
\label{Section5.1}

This section presents the experimental results using the ADAGE framework for multimodal post-flood water extent mapping. This case study utilized Synthetic Aperture Radar (SAR) and Multispectral Imaging (MSI) data and investigated explanation alignment in cloud-covered areas, where domain knowledge has been established in previous studies. Evaluations were conducted only in regions identified as cloud-covered based on the cloud labels provided in the C2S-MS Floods dataset.

In this experiment, the ADAGE framework was configured as defined in Table \ref{tab:tab3}. Experiments were repeated 10 times using U-Net and U-Net++ with a ResNet-50 encoder, SegFormer with a MiT-B3 encoder, and two recent state-of-the-art models, PyramidMamba and SMAGNet. Based on the experimental results, the mean and standard deviation of four performance metrics and the alignment scores regarding the two reference explanations, $\text{RE}_\text{case1-1}$ and $\text{RE}_\text{case1-2}$ defined in Table \ref{tab:tab3}, are presented in Table \ref{tab:tab5}. 

\begin{table}[]
\caption{Performance metrics and alignment scores of U-Net, U-Net++, SegFormer, PyramidMamba, and SMAGNet models using the ADAGE framework in multimodal post-flood water extent mapping in cloud-covered areas.}
\label{tab:tab5}
\resizebox{0.75\textwidth}{!}{%
\begin{tabular}{@{}ccccccc@{}}
\toprule
\multirow{2}{*}{\textbf{Model}} &
\multirow{2}{*}{\textbf{IoU}} &
\multirow{2}{*}{\textbf{Precision}} &
\multirow{2}{*}{\textbf{Recall}} &
\multirow{2}{*}{\textbf{F1-Score}} &
\multicolumn{2}{c}{\textbf{Alignment Score}} \\ \cmidrule(l){6-7}
& & & & & $\text{RE}_{\text{case1-1}}$ & $\text{RE}_{\text{case1-2}}$ \\ \midrule \midrule

U-Net &
\begin{tabular}[c]{@{}c@{}}84.00\\ (±0.75)\end{tabular} &
\begin{tabular}[c]{@{}c@{}}92.98\\ (±1.40)\end{tabular} &
\begin{tabular}[c]{@{}c@{}}89.72\\ (±1.50)\end{tabular} &
\begin{tabular}[c]{@{}c@{}}91.30\\ (±0.45)\end{tabular} &
\begin{tabular}[c]{@{}c@{}}78.35\\ (±13.35)\end{tabular} &
\begin{tabular}[c]{@{}c@{}}98.21\\ (±1.70)\end{tabular} \\ \midrule

U-Net++ &
\begin{tabular}[c]{@{}c@{}}84.35\\ (±0.81)\end{tabular} &
\begin{tabular}[c]{@{}c@{}}93.04\\ (±1.77)\end{tabular} &
\begin{tabular}[c]{@{}c@{}}90.07\\ (±1.24)\end{tabular} &
\begin{tabular}[c]{@{}c@{}}91.51\\ (±0.47)\end{tabular} &
\begin{tabular}[c]{@{}c@{}}77.93\\ (±15.45)\end{tabular} &
\begin{tabular}[c]{@{}c@{}}99.19\\ (±0.85)\end{tabular} \\ \midrule

SegFormer &
\begin{tabular}[c]{@{}c@{}}83.57\\ (±1.22)\end{tabular} &
\begin{tabular}[c]{@{}c@{}}93.32\\ (±2.35)\end{tabular} &
\begin{tabular}[c]{@{}c@{}}88.96\\ (±2.15)\end{tabular} &
\begin{tabular}[c]{@{}c@{}}91.04\\ (±0.73)\end{tabular} &
\begin{tabular}[c]{@{}c@{}}54.65\\ (±24.41)\end{tabular} &
\begin{tabular}[c]{@{}c@{}}98.80\\ (±1.41)\end{tabular} \\ \midrule

PyramidMamba &
\begin{tabular}[c]{@{}c@{}}85.70\\ (±1.01)\end{tabular} &
\begin{tabular}[c]{@{}c@{}}94.38\\ (±1.43)\end{tabular} &
\begin{tabular}[c]{@{}c@{}}90.32\\ (±0.73)\end{tabular} &
\begin{tabular}[c]{@{}c@{}}92.30\\ (±0.59)\end{tabular} &
\begin{tabular}[c]{@{}c@{}}84.22\\ (±11.20)\end{tabular} &
\begin{tabular}[c]{@{}c@{}}95.72\\ (±2.25)\end{tabular} \\ \midrule

SMAGNet &
\begin{tabular}[c]{@{}c@{}}85.33\\ (±0.80)\end{tabular} &
\begin{tabular}[c]{@{}c@{}}93.97\\ (±0.75)\end{tabular} &
\begin{tabular}[c]{@{}c@{}}90.28\\ (±1.04)\end{tabular} &
\begin{tabular}[c]{@{}c@{}}92.08\\ (±0.47)\end{tabular} &
\begin{tabular}[c]{@{}c@{}}90.91\\ (±10.25)\end{tabular} &
\begin{tabular}[c]{@{}c@{}}99.33\\ (±0.84)\end{tabular} \\ 

\bottomrule
\end{tabular}%
}
\end{table}

In Table \ref{tab:tab5},  flood segmentation performance was comparable across all five models in cloud-covered areas, with the recently proposed PyramidMamba and SMAGNet achieving slightly better performance than U-Net, U-Net++, and SegFormer. Importantly, when examining the alignment scores, distinct patterns were observed between the alignment scores for $\text{RE}_\text{case1-1}$ and $\text{RE}_\text{case1-2}$. While the alignment scores for $\text{RE}_\text{case1-1}$ varied considerably across models, ranging from 54.65\% to 90.91\%, the alignment scores for $\text{RE}_\text{case1-2}$ consistently exceeded 95\% for all five models. Among the five models, SMAGNet achieved the highest alignment scores for both reference explanations, with an alignment score of 90.91\% for $\text{RE}_\text{case1-1}$ and 99.33\% for $\text{RE}_\text{case1-2}$. These results indicate that, although the models achieved comparable flood segmentation performance, SMAGNet exhibited the strongest consistency with the domain knowledge. Furthermore, the high alignment scores with $\text{RE}_\text{case1-2}$ suggest that all models primarily relied on the complementary use of SAR and NIR information under thin cloud cover, whereas the larger variation in $\text{RE}_\text{case1-1}$ implies that the extent to which the models relied on SAR alone differed depending on the model architecture. 

\begin{figure}[]
  \centering
  \includegraphics[width=1.0 \linewidth]{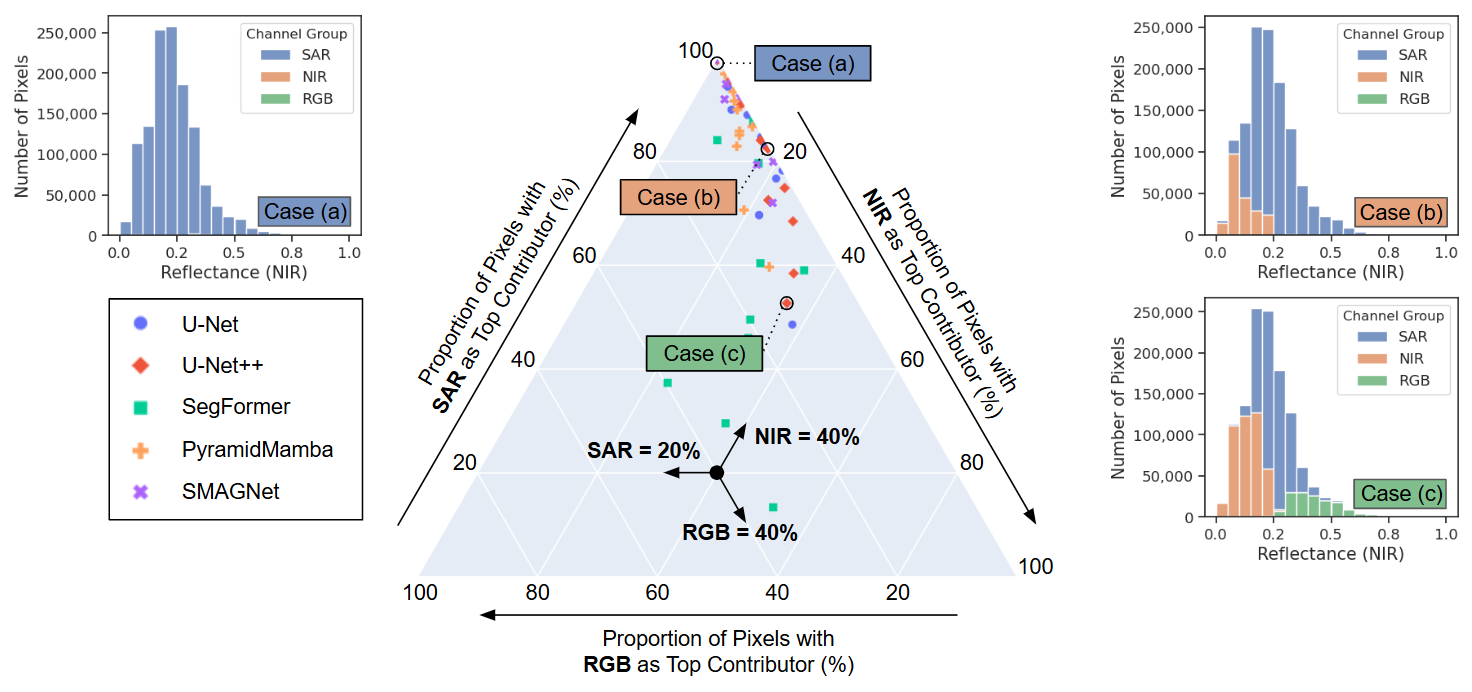}
  \caption{Ternary plot visualizing the proportion of channel groups that contributed the most to true positive water pixel predictions under cloud-covered conditions, with stacked histograms for the three cases showing the number of true positive water pixels according to NIR reflectance.}
  \label{fig2}
\end{figure}

To support a more in-depth analysis of the alignment scores for each reference explanation, Fig. \ref{fig2} shows a ternary plot visualizing the proportion of the channel groups that made the largest contribution to correctly predicted water pixels in cloud-covered areas across the 10 repeated experiments using five different models.
In Fig. \ref{fig2}, the three axes correspond to the proportion of channel groups (SAR, RGB, and NIR) that contributed the most to water pixel predictions, with values ranging from 0\% to 100\%. The blue circular, red diamond, green square, orange cross, and purple star markers represent the experiments using U-Net, U-Net++, SegFormer, PyramidMamba, and SMAGNet, respectively. Each point represents the result of a single independently trained model, and its position is determined by the proportion of each channel group that made the largest contribution to true positive water pixel predictions.

As shown in Fig. \ref{fig2}, the five models exhibit distinct distribution patterns in the ternary plot, indicating that different model architectures exploit the three channel groups differently for water pixel prediction under cloud-covered conditions. To investigate the characteristics of models positioned on the ternary plot, three representative models, cases (a), (b), and (c), were selected. Notably, cases (a), (b), and (c) show models with different proportions of channel groups that contribute most to true positive water pixel predictions under cloudy conditions, even when trained under the same configurations. As in case (a), models located near the apex of the ternary plot indicate that the SAR channel group contributes the most to nearly all true positive water pixel predictions. Additionally, in Fig. \ref{fig2}, moving from case (a) to case (b), there is a trend in which the proportion of the NIR channel group as the top contributor increases up to 20\%, while the proportion of the SAR channel group as the top contributor decreases to 80\%. Moreover, from case (b) to case (c), the proportions of the RGB and NIR channel groups contributing the most to true positive water pixel predictions gradually increase, while the proportion of the SAR channel group contributing the most decreases. In particular, the model corresponding to case (c) exhibits a relatively higher proportion (around 10\% or more) of the RGB channel group contributing the most compared to other cases.

\begin{figure}[]
  \centering
  \includegraphics[width=1.0 \linewidth]{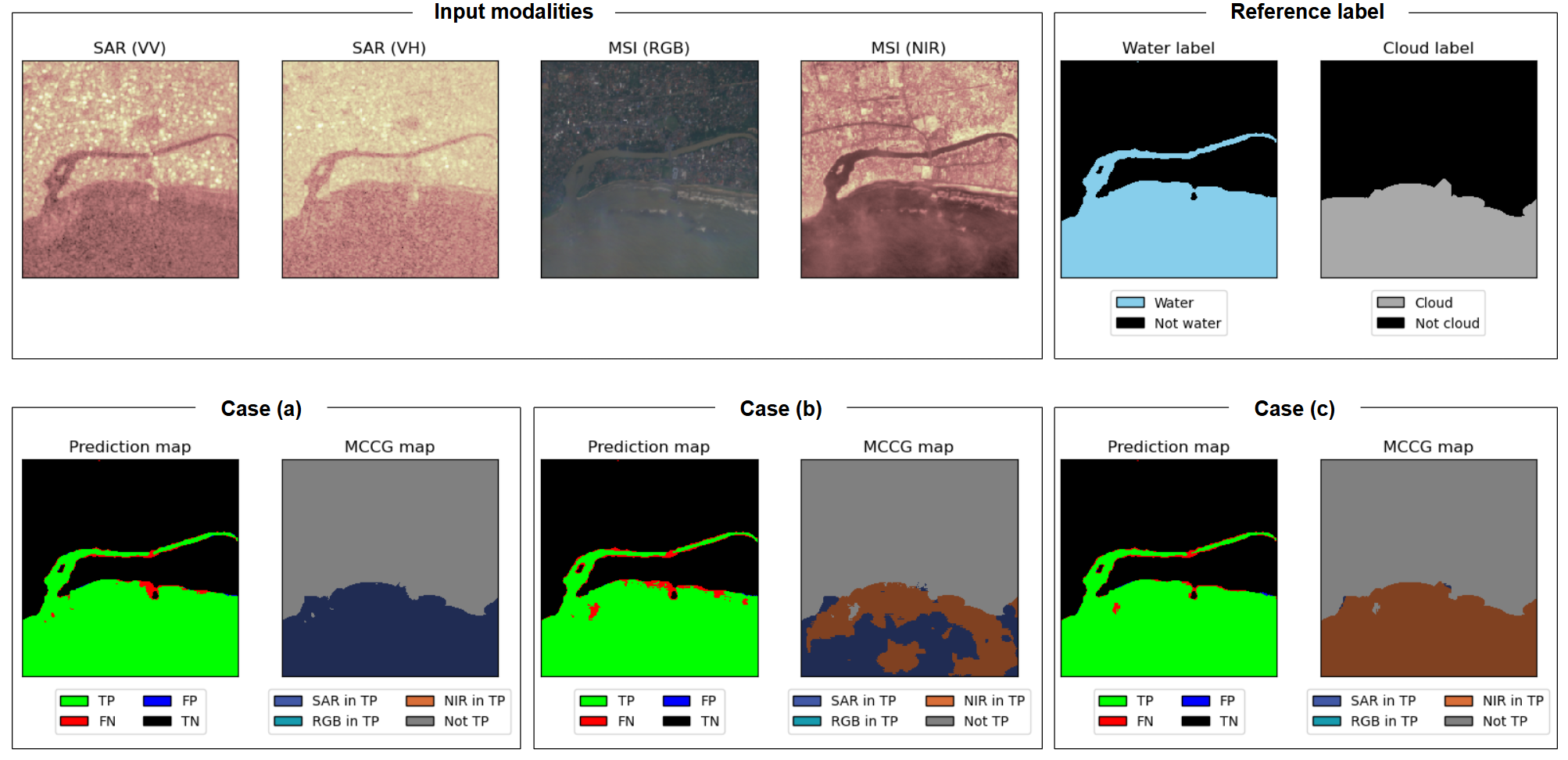}
  \caption{Visualization of prediction maps and corresponding Most Contributed Channel Group (MCCG) maps for cases (a), (b), and (c) for a single individual sample.}
  \label{fig3}
\end{figure}

\begin{figure}[]
  \centering
  \includegraphics[width=1.0 \linewidth]{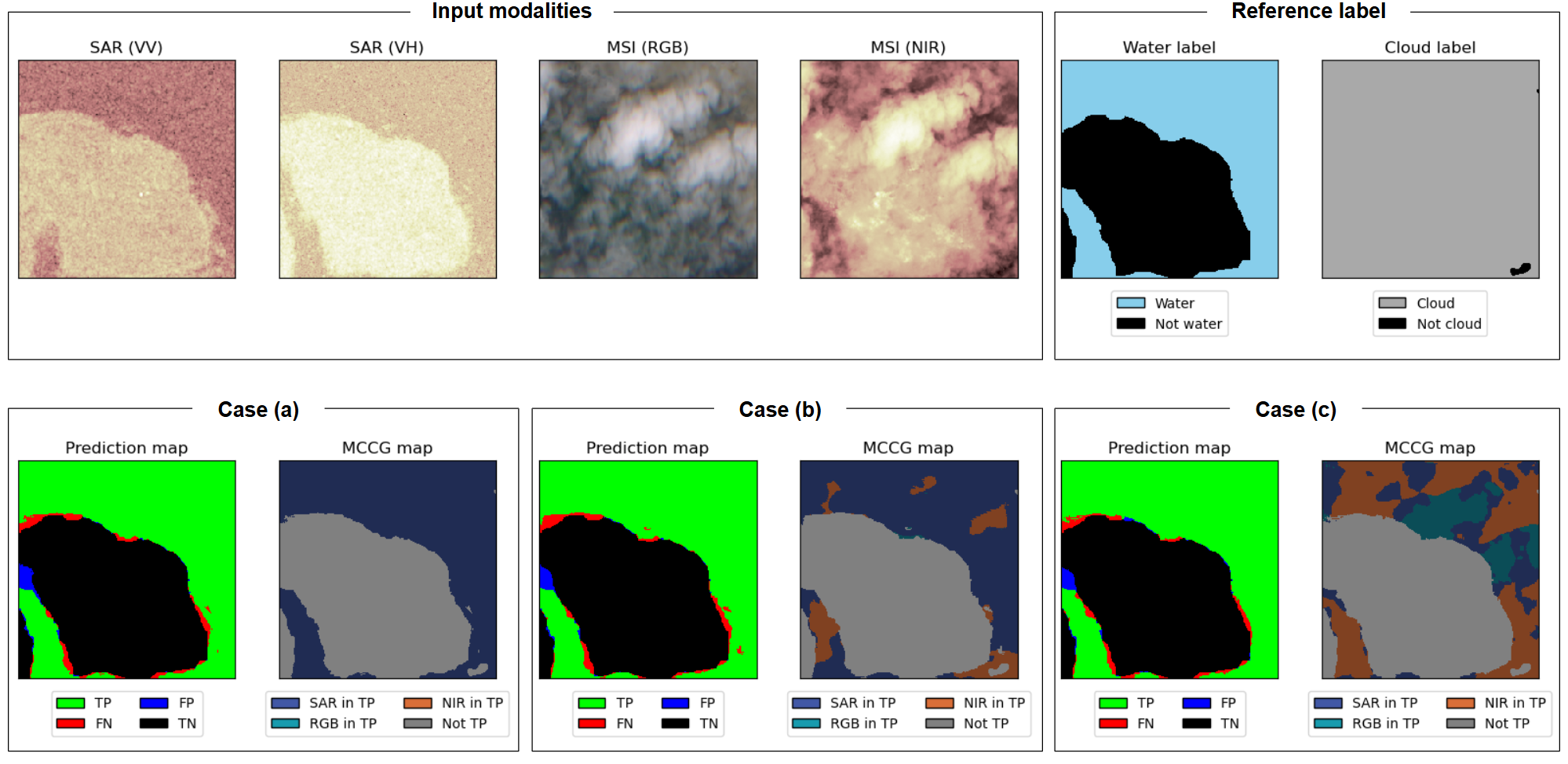}
  \caption{Visualization of prediction maps and corresponding Most Contributed Channel Group (MCCG) maps for cases (a), (b), and (c) for another individual sample.}
  \label{fig4}
\end{figure}

For each case, stacked histograms were plotted, with the x-axis representing NIR reflectance and the y-axis representing the number of true positive water pixels to which each channel group contributed the most. The histogram of case (a) showed that the SAR channel group contributed the most for the majority of pixels across the entire range of NIR reflectance values. On the other hand, the histogram of case (b) illustrated that the NIR channel group, along with the SAR channel group, made the largest contribution within the NIR reflectance range below 0.2. Since this experiment was conducted under cloud conditions based on cloud labels, we interpreted low NIR reflectance values as corresponding to relatively thin cloud conditions, whereas high NIR reflectance values indicated thick cloud conditions. Finally, in the histogram of case (c), within the low NIR reflectance range (below 0.2), a pattern similar to case (b) was observed, whereas for NIR reflectance values above 0.3, the RGB channel group tended to make the largest contribution to a portion of the true positive water pixels.

The characteristics of case (a), (b), and (c) regarding channel groups shown in the ternary plot and stacked histogram are also consistently demonstrated in the individual samples presented in Figs. \ref{fig3} and \ref{fig4}. Although Figs. \ref{fig3} and \ref{fig4} present visually similar prediction maps across cases (a), (b), and (c) (independent training runs using the same model architecture, configuration, and training data), each case's Most Contributed Channel Group (MCCG) map, which illustrates the channel group that made the largest contribution to the prediction of each true positive water pixel under cloudy conditions, shows different characteristics. Specifically, in case (a), the SAR channel group contributed the most to the prediction of the majority of true positive water pixels. In contrast, as the cases shift to (b) and (c), the number of pixels to which the NIR and RGB channel groups make the largest contribution increases. Notably, in the MCCG maps shown in Fig. \ref{fig3}, the channel group with the largest contribution is either the NIR or the SAR channel group, depending on how the NIR reflectance is affected by cloud presence. The MCCG maps for case (c) in Fig. \ref{fig4} exhibit a tendency for the RGB channel group to make the largest contribution to the prediction of true positive water pixels under cloud conditions, particularly in regions with thick cloud cover. 

\subsection{Case Study 2: SAR-based Open and Urban Area Flood Mapping}
\label{Section5.2}

\subsubsection{Flooded Open Area Mapping}
\label{Section5.2.1}

This section presents the experimental results of the ADAGE framework for flooded open and flooded urban area mapping using pre- and post-flood SAR data.
The DLSS-RS models were trained using rule-based labeled data, and the channel groups used for the labeling process were regarded as reference explanations. The ADAGE framework configuration summarized in Table \ref{tab:tab4} was used throughout the experiments. Each experiment was repeated 10 times using the same three DLSS-RS models as in the previous experiments: U-Net and U-Net++ with a ResNet-50 encoder, and SegFormer with a MiT-B3 encoder. For flooded open area mapping, Table \ref{tab:tab6} reports the mean and standard deviation of four performance metrics, along with the alignment score for the reference explanation $\text{RE}_\text{case2-1}$, which states that flooded open areas are annotated using the SAR intensity channel group.

In SAR-based flooded open area mapping, segmentation performance varied across the three test regions. Specifically, U-Net and U-Net++ achieved high IoU scores in the Weihui (84.12\% and 83.61\%) and Jubba (81.03\% and 82.16\%) regions. In both regions, the IoU standard deviations remained below 2.1\%, indicating stable model performance across repeated experiments. In contrast, lower IoU scores were observed in the NovaKakhovka region (43.44\% for U-Net, 43.17\% for U-Net++, and 41.33\% for SegFormer), accompanied by larger standard deviations, particularly for U-Net++. Despite regional differences, U-Net and U-Net++ exhibited comparable overall performance and consistently outperformed SegFormer in terms of IoU and F1-score. In contrast, SegFormer consistently achieved slightly lower segmentation performance than U-Net and U-Net++ across all test regions.

\begin{table}[t]
\caption{Performance metrics and alignment score for the flooded open area using the ADAGE framework with U-Net, U-Net++, and SegFormer in SAR-based flood mapping.}
\label{tab:tab6}
\resizebox{0.75\textwidth}{!}{%
\begin{tabular}{@{}ccccccc@{}}
\toprule
Test Region          & Model   & IoU                                                       & Precision                                                 & Recall                                                    & F1-Score                                                  & \begin{tabular}[c]{@{}c@{}}Alignment Score: \\ $\text{RE}_\text{case2-1}$ \end{tabular} \\ \midrule \midrule
Weihui               & U-Net   & \begin{tabular}[c]{@{}c@{}}84.12 \\ (±0.26)\end{tabular} & \begin{tabular}[c]{@{}c@{}}92.20 \\ (±1.06)\end{tabular} & \begin{tabular}[c]{@{}c@{}}90.58 \\ (±1.09)\end{tabular} & \begin{tabular}[c]{@{}c@{}}91.37 \\ (±0.15)\end{tabular} & \begin{tabular}[c]{@{}c@{}}99.94 \\ (±0.17)\end{tabular} \\ \cmidrule(l){2-7}
                     & U-Net++ & \begin{tabular}[c]{@{}c@{}}83.61 \\ (±0.53)\end{tabular} & \begin{tabular}[c]{@{}c@{}}92.88 \\ (±1.33)\end{tabular} & \begin{tabular}[c]{@{}c@{}}89.38 \\ (±1.62)\end{tabular} & \begin{tabular}[c]{@{}c@{}}91.07 \\ (±0.31)\end{tabular} & \begin{tabular}[c]{@{}c@{}}99.94 \\ (±0.16)\end{tabular} \\ \cmidrule(l){2-7}
                     & SegFormer & \begin{tabular}[c]{@{}c@{}}80.32 \\ (±0.48)\end{tabular} & \begin{tabular}[c]{@{}c@{}}90.57 \\ (±1.30)\end{tabular} & \begin{tabular}[c]{@{}c@{}}87.68 \\ (±1.34)\end{tabular} & \begin{tabular}[c]{@{}c@{}}89.09 \\ (±0.30)\end{tabular} & \begin{tabular}[c]{@{}c@{}}100.00 \\ (±0.00)\end{tabular} \\ \midrule

Jubba                & U-Net   & \begin{tabular}[c]{@{}c@{}}81.03 \\ (±2.05)\end{tabular} & \begin{tabular}[c]{@{}c@{}}82.34 \\ (±2.47)\end{tabular} & \begin{tabular}[c]{@{}c@{}}98.11 \\ (±0.54)\end{tabular} & \begin{tabular}[c]{@{}c@{}}89.51 \\ (±1.26)\end{tabular} & \begin{tabular}[c]{@{}c@{}}99.54 \\ (±1.44)\end{tabular} \\ \cmidrule(l){2-7}
                     & U-Net++ & \begin{tabular}[c]{@{}c@{}}82.16 \\ (±1.79)\end{tabular} & \begin{tabular}[c]{@{}c@{}}83.62 \\ (±2.14)\end{tabular} & \begin{tabular}[c]{@{}c@{}}97.95 \\ (±0.61)\end{tabular} & \begin{tabular}[c]{@{}c@{}}90.20 \\ (±1.09)\end{tabular} & \begin{tabular}[c]{@{}c@{}}99.96 \\ (±0.11)\end{tabular} \\ \cmidrule(l){2-7}
                     & SegFormer & \begin{tabular}[c]{@{}c@{}}79.95 \\ (±2.05)\end{tabular} & \begin{tabular}[c]{@{}c@{}}83.01 \\ (±3.38)\end{tabular} & \begin{tabular}[c]{@{}c@{}}95.72 \\ (±1.73)\end{tabular} & \begin{tabular}[c]{@{}c@{}}88.85 \\ (±1.27)\end{tabular} & \begin{tabular}[c]{@{}c@{}}100.00 \\ (±0.00)\end{tabular} \\ \midrule

NovaKakhovka         & U-Net   & \begin{tabular}[c]{@{}c@{}}43.44 \\ (±2.66)\end{tabular} & \begin{tabular}[c]{@{}c@{}}51.53 \\ (±5.79)\end{tabular} & \begin{tabular}[c]{@{}c@{}}74.69 \\ (±5.76)\end{tabular} & \begin{tabular}[c]{@{}c@{}}60.53 \\ (±2.61)\end{tabular} & \begin{tabular}[c]{@{}c@{}}99.33 \\ (±0.50)\end{tabular} \\ \cmidrule(l){2-7}
                     & U-Net++ & \begin{tabular}[c]{@{}c@{}}43.17 \\ (±4.10)\end{tabular} & \begin{tabular}[c]{@{}c@{}}52.43 \\ (±8.08)\end{tabular} & \begin{tabular}[c]{@{}c@{}}72.71 \\ (±6.56)\end{tabular} & \begin{tabular}[c]{@{}c@{}}60.20 \\ (±3.93)\end{tabular} & \begin{tabular}[c]{@{}c@{}}99.45 \\ (±0.44)\end{tabular} \\ \cmidrule(l){2-7}
                     & SegFormer & \begin{tabular}[c]{@{}c@{}}41.33 \\ (±2.48)\end{tabular} & \begin{tabular}[c]{@{}c@{}}56.09 \\ (±4.56)\end{tabular} & \begin{tabular}[c]{@{}c@{}}61.60 \\ (±5.16)\end{tabular} & \begin{tabular}[c]{@{}c@{}}58.45 \\ (±2.46)\end{tabular} & \begin{tabular}[c]{@{}c@{}}99.70 \\ (±0.28)\end{tabular} \\ \bottomrule
\end{tabular}%
}
\end{table}


\begin{table}[!ht]
\caption{Performance metrics and alignment score for the flooded urban area using the ADAGE framework with U-Net, U-Net++, and SegFormer in SAR-based flood mapping.}
\label{tab:tab7}
\resizebox{0.75\textwidth}{!}{%
\begin{tabular}{@{}ccccccc@{}}
\toprule
Test Region          & Model   & IoU                                                       & Precision                                                 & Recall                                                    & F1-Score                                                  & \begin{tabular}[c]{@{}c@{}}Alignment Score:\\ $\text{RE}_\text{case2-2}$\end{tabular} \\ \midrule \midrule

Weihui
& U-Net
& \begin{tabular}[c]{@{}c@{}}27.44 \\ (±3.95)\end{tabular}
& \begin{tabular}[c]{@{}c@{}}68.61 \\ (±4.07)\end{tabular}
& \begin{tabular}[c]{@{}c@{}}31.79 \\ (±6.05)\end{tabular}
& \begin{tabular}[c]{@{}c@{}}42.93 \\ (±4.91)\end{tabular}
& \begin{tabular}[c]{@{}c@{}}85.17 \\ (±13.04)\end{tabular} \\ \cmidrule(l){2-7}

& U-Net++
& \begin{tabular}[c]{@{}c@{}}24.18 \\ (±4.82)\end{tabular}
& \begin{tabular}[c]{@{}c@{}}71.79 \\ (±4.31)\end{tabular}
& \begin{tabular}[c]{@{}c@{}}27.07 \\ (±6.28)\end{tabular}
& \begin{tabular}[c]{@{}c@{}}38.72 \\ (±6.44)\end{tabular}
& \begin{tabular}[c]{@{}c@{}}88.22 \\ (±7.38)\end{tabular} \\ \cmidrule(l){2-7}

& SegFormer
& \begin{tabular}[c]{@{}c@{}}21.14 \\ (±3.89)\end{tabular}
& \begin{tabular}[c]{@{}c@{}}64.34 \\ (±4.05)\end{tabular}
& \begin{tabular}[c]{@{}c@{}}24.22 \\ (±5.31)\end{tabular}
& \begin{tabular}[c]{@{}c@{}}34.75 \\ (±5.35)\end{tabular}
& \begin{tabular}[c]{@{}c@{}}79.84 \\ (±6.45)\end{tabular} \\ \midrule

Jubba
& U-Net
& \begin{tabular}[c]{@{}c@{}}53.01 \\ (±2.49)\end{tabular}
& \begin{tabular}[c]{@{}c@{}}77.14 \\ (±1.90)\end{tabular}
& \begin{tabular}[c]{@{}c@{}}63.03 \\ (±4.28)\end{tabular}
& \begin{tabular}[c]{@{}c@{}}69.26 \\ (±2.13)\end{tabular}
& \begin{tabular}[c]{@{}c@{}}90.54 \\ (±9.36)\end{tabular} \\ \cmidrule(l){2-7}

& U-Net++
& \begin{tabular}[c]{@{}c@{}}51.78 \\ (±3.08)\end{tabular}
& \begin{tabular}[c]{@{}c@{}}78.61 \\ (±1.73)\end{tabular}
& \begin{tabular}[c]{@{}c@{}}60.43 \\ (±4.93)\end{tabular}
& \begin{tabular}[c]{@{}c@{}}68.18 \\ (±2.69)\end{tabular}
& \begin{tabular}[c]{@{}c@{}}91.41 \\ (±9.23)\end{tabular} \\ \cmidrule(l){2-7}

& SegFormer
& \begin{tabular}[c]{@{}c@{}}52.40 \\ (±1.36)\end{tabular}
& \begin{tabular}[c]{@{}c@{}}70.53 \\ (±3.50)\end{tabular}
& \begin{tabular}[c]{@{}c@{}}67.37 \\ (±3.86)\end{tabular}
& \begin{tabular}[c]{@{}c@{}}68.76 \\ (±1.17)\end{tabular}
& \begin{tabular}[c]{@{}c@{}}82.95 \\ (±9.01)\end{tabular} \\ \midrule

NovaKakhovka
& U-Net
& \begin{tabular}[c]{@{}c@{}}48.39 \\ (±2.58)\end{tabular}
& \begin{tabular}[c]{@{}c@{}}72.12 \\ (±3.19)\end{tabular}
& \begin{tabular}[c]{@{}c@{}}59.94 \\ (±5.87)\end{tabular}
& \begin{tabular}[c]{@{}c@{}}65.18 \\ (±2.37)\end{tabular}
& \begin{tabular}[c]{@{}c@{}}93.42 \\ (±6.60)\end{tabular} \\ \cmidrule(l){2-7}

& U-Net++
& \begin{tabular}[c]{@{}c@{}}46.94 \\ (±3.09)\end{tabular}
& \begin{tabular}[c]{@{}c@{}}72.79 \\ (±3.13)\end{tabular}
& \begin{tabular}[c]{@{}c@{}}57.37 \\ (±6.33)\end{tabular}
& \begin{tabular}[c]{@{}c@{}}63.84 \\ (±2.88)\end{tabular}
& \begin{tabular}[c]{@{}c@{}}92.42 \\ (±7.58)\end{tabular} \\ \cmidrule(l){2-7}

& SegFormer
& \begin{tabular}[c]{@{}c@{}}37.96 \\ (±4.86)\end{tabular}
& \begin{tabular}[c]{@{}c@{}}72.12 \\ (±2.39)\end{tabular}
& \begin{tabular}[c]{@{}c@{}}44.83 \\ (±7.07)\end{tabular}
& \begin{tabular}[c]{@{}c@{}}54.86 \\ (±5.43)\end{tabular}
& \begin{tabular}[c]{@{}c@{}}92.83 \\ (±3.32)\end{tabular} \\

\bottomrule
\end{tabular}%
}
\end{table}

As described in Section \ref{Section4.2}, the flooded open (FO) class in the training data was labeled using the pre- and post-flood VH and VV intensity channels. In contrast, the labels in the test data were manually annotated by human experts. For true positive FO pixels across all three test regions, the alignment scores were consistently above 99\%. This demonstrates that the model explanations remain highly aligned with the reference explanation derived from the labeling rule, even when evaluated on test data independently annotated by human experts.

\subsubsection{Flooded Urban Area Mapping}
\label{Section5.2.2}

Table \ref{tab:tab7} summarizes the mean and standard deviation of four performance metrics for flooded urban area mapping, along with the alignment score for the reference explanation $\text{RE}_\text{case2-2}$, which states that flooded urban areas are annotated based on SAR interferometric coherence values and the WSF2019 dataset. In SAR-based flooded urban area mapping, segmentation performance varied across the three test regions. U-Net and U-Net++ achieved comparable IoU scores in the Jubba and NovaKakhovka regions, while SegFormer achieved a comparable IoU in Jubba but noticeably lower performance in NovaKakhovka. In both regions, the IoU standard deviations remained below 5\%, indicating stable performance across repeated training runs. In contrast, all three models exhibited substantially lower IoU values in Weihui, with 27.44\%, 24.18\%, and 21.14\% for U-Net, U-Net++, and SegFormer, respectively. Overall, U-Net and U-Net++ demonstrated consistently comparable performance across all test regions, whereas SegFormer underperformed, with the largest performance degradation observed in the NovaKakhovka region.


As presented in Section \ref{Section4.2}, the flooded urban (FU) class in the training data was labeled using the coherence of the pre- and post-flood VH and VV bands in combination with WSF 2019 data, but the labels in the test data were manually annotated by human experts. For true positive FU pixels, the alignment scores with the reference explanation ranged from 79.84\% (SegFormer model prediction in Weihui) to 93.42\% (U-Net model prediction in NovaKakhovka), which were consistently lower than those observed for flooded open areas. This indicates a lower alignment between the reference explanation $\text{RE}_\text{case2-2}$, which is derived from the labeling rule for the FU class, and 
the model explanations evaluated on the expert-annotated test data.

\begin{figure}[]
  \centering
  \includegraphics[width=1.0 \linewidth]{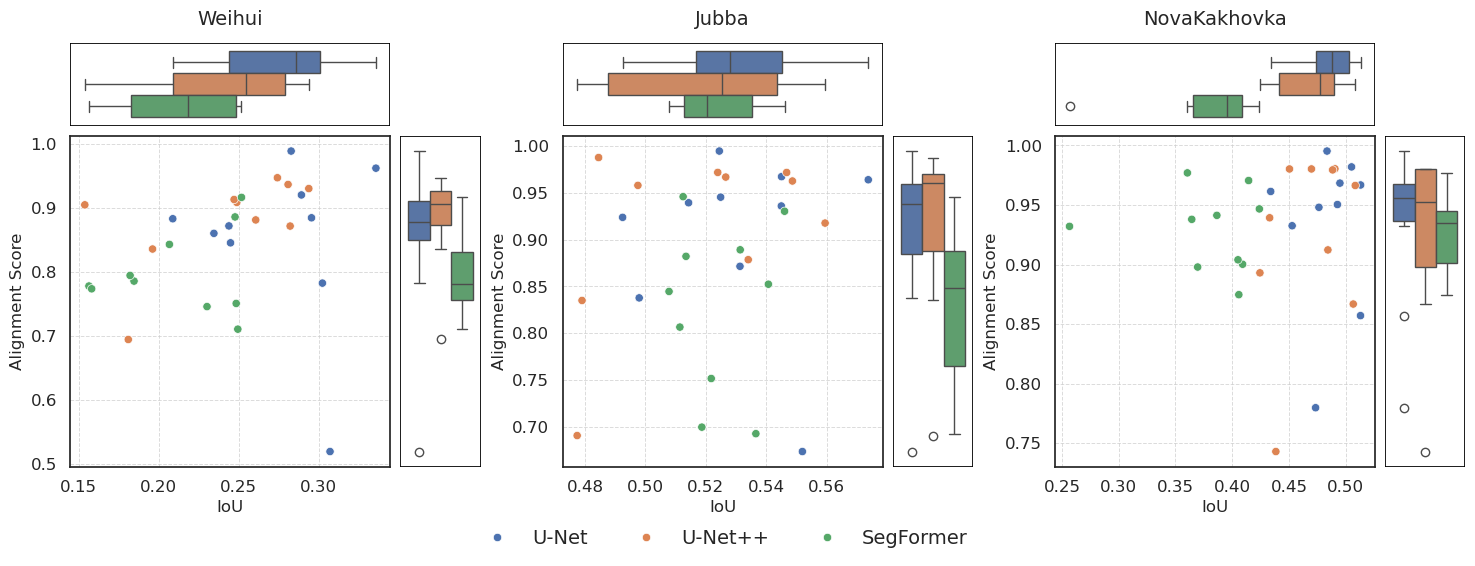}
  \caption{Scatterplot showing the relationship between IoU and alignment score in flooded urban area mapping.}
  \label{fig5}
\end{figure}

Fig. \ref{fig5} shows a scatter plot of IoU and alignment score to further examine their relationship in flooded urban area mapping. In Fig. \ref{fig5}, no clear correlation between IoU and alignment score was observed across the three regions, suggesting that the two metrics may capture different aspects of model behavior. Therefore, when deployment requires a high degree of alignment with a specific reference explanation, both IoU and the alignment score should be considered during model selection. The procedure for using alignment scores in the ADAGE framework is further discussed in the Discussion section.

\section{Discussion}
\label{Section6}

Through the two flood mapping case studies in Section \ref{Section5}, the ADAGE framework demonstrated the capability to quantitatively evaluate the alignment between model explanations and reference explanations derived from domain knowledge. The alignment score provides a systematic metric for assessing how closely model explanations align with scientifically grounded reference explanations. High alignment scores indicate strong agreement with established knowledge, whereas low scores suggest that models rely on patterns inconsistent with the reference domain knowledge.
 
In Section \ref{Section5.1}, one important observation is that the degree of alignment between the model explanations and the reference explanation varies across model architectures, even when their segmentation performance is comparable. As shown in Table \ref{tab:tab5}, U-Net and U-Net++ achieved similar alignment scores for both $\text{RE}_\text{case1-1}$ and $\text{RE}_\text{case1-2}$, whereas SegFormer exhibited substantially lower alignment with $\text{RE}_\text{case1-1}$ despite achieving comparable IoU and F1-score. In contrast, PyramidMamba and SMAGNet showed higher alignment with $\text{RE}_\text{case1-1}$, with SMAGNet achieving the highest alignment among all evaluated models. These results suggest that the alignment between the model explanations and the reference explanation is not solely determined by segmentation performance or broad architectural categories, but also depends on the feature utilization strategy learned by each model. Therefore, the alignment score provides complementary information to conventional performance metrics by quantifying how closely the model explanations are aligned with the reference explanation derived from domain knowledge.

Furthermore, the standard deviations of the alignment scores indicate that explanation consistency can vary among independently trained instances of the same model architecture, even when the training configuration is identical. While the segmentation performance metrics exhibited relatively small standard deviations, the alignment scores exhibited noticeably larger standard deviations, particularly for $\text{RE}_\text{case1-1}$. This observation suggests that the stochastic optimization process, including random initialization and mini-batch sampling, may lead identical architectures to converge to different feature utilization strategies while maintaining similar prediction performance. Consequently, models with comparable segmentation accuracy may exhibit substantially different levels of alignment with the same reference explanation. 
These findings highlight the importance of considering the variability of alignment score across independently trained models of the same architecture when interpreting the explanation characteristics of the architecture, particularly in applications where alignment with reference explanations derived from domain knowledge is an important criterion for model selection and deployment.

For example, Fig. \ref{fig3} shows that three independently trained instances of the same model architecture produce similar prediction results under thin cloud conditions, while exhibiting different channel group contribution patterns.
This variability is further illustrated in Fig. \ref{fig4}(c), where the model exhibits the highest contribution from the RGB channel group within regions covered by thick clouds, even though RGB information is unlikely to be directly informative for flood detection in the presence of thick cloud cover. This pattern may indicate a reliance on spurious correlations arising from the frequent co-occurrence of thick cloud cover and flood events, potentially reflecting shortcut learning rather than physically meaningful flood-related evidence.

A similar interpretation can be drawn in Section \ref{Section5.2}. In the flooded open area mapping, the alignment scores with the reference explanation $\text{RE}_\text{case2-1}$ exceeded 99\% across all three test regions, with low standard deviations. This indicates a strong alignment between the model explanations and the reference explanation derived from domain knowledge, suggesting that the models' decision-making process aligns well with established scientific understanding that SAR intensity is effective for detecting flooded open areas. On the other hand, in the flooded urban area mapping scenario, the alignment score with $\text{RE}_\text{case2-2}$ was relatively lower than that with $\text{RE}_\text{case2-1}$. This indicates that during the models' inference, channel groups not included in the reference explanation $\text{RE}_\text{case2-2}$ had relatively greater contributions. In other words, for flooded open area mapping, models trained with rule-based labeling using SAR intensity inferred in a manner aligned with the training data labeling rule even on manually labeled test data. In contrast, for flooded urban area mapping, models trained with rule-based labels derived from SAR interferometric coherence and WSF2019 produced explanations that showed lower alignment with the corresponding reference explanation, indicating that the models relied on feature utilization patterns that were only partially aligned with the labeling rule when evaluated on manually labeled test data. In addition, Fig. \ref{fig5} shows that the alignment score captures a different aspect of model behavior from IoU, indicating that both metrics can be considered together for model selection.

In summary, the ADAGE framework provides a systematic approach to evaluate the alignment score between model explanations and reference explanations. High alignment scores, such as those observed in $\text{RE}_\text{case1-2}$ and $\text{RE}_\text{case2-1}$, indicate that the models' decision-making processes closely align with domain knowledge. In other words, the models closely follow the reference explanation in their correct predictions. On the other hand, low alignment scores, as in $\text{RE}_\text{case1-1}$ and $\text{RE}_\text{case2-2}$, indicate diverse strategies and potential misalignment, which can be further explored. This approach allows domain experts to identify models that may not follow established knowledge and provides an additional criterion for model selection alongside existing performance metrics, thereby supporting more trustworthy model selection for operational flood mapping.

\begin{figure}[]
  \centering
  \includegraphics[width=1.0 \linewidth]{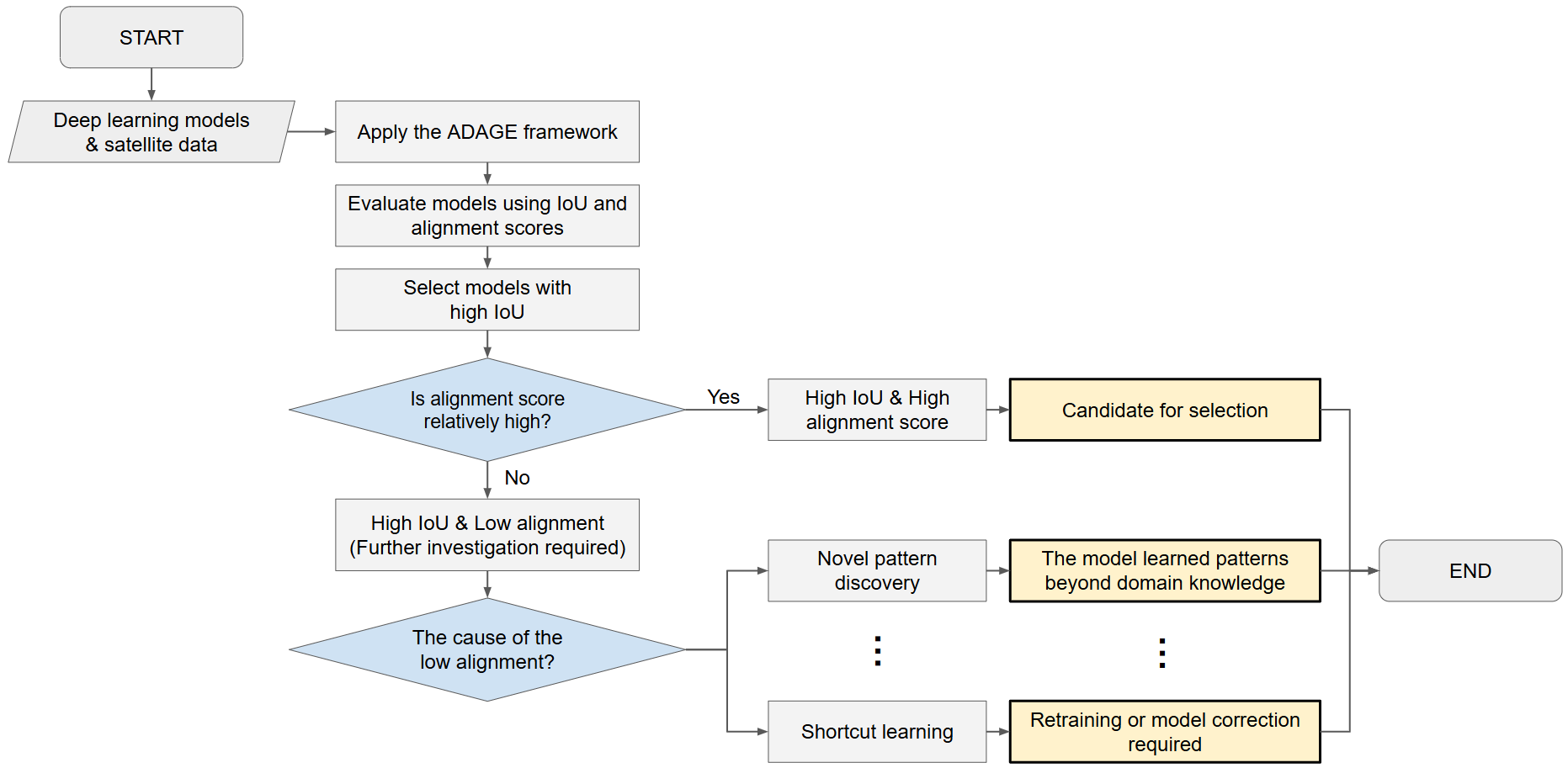}
  \caption{Workflow for applying the ADAGE framework to DLSS-RS models.}
  \label{fig6}
\end{figure}

Based on these findings and inspired by \citep{jiang2024interpretable}, Fig. \ref{fig6} illustrates a workflow for applying and interpreting the ADAGE framework to select DLSS-RS models aligned with domain knowledge in real-world deployment. After the DLSS-RS models are trained, the ADAGE framework provides a domain knowledge alignment score that indicates how well the model's predictions align with established domain knowledge, in addition to the IoU score commonly used to evaluate the model's predictive accuracy. While model selection typically relies on predictive accuracy metrics such as the IoU, the ADAGE framework enables the evaluation of a model’s trustworthiness through this additional alignment metric. Models with both high predictive performance and strong alignment scores are ideal candidates for real-world deployment. In contrast, models with high IoU but low alignment scores should be treated with caution, as low alignment scores provide an important warning to domain experts that the models may not be consistent with reference explanations derived from domain knowledge. In such cases, further investigation is required to determine the cause. Low alignment may result from novel pattern discovery, where the model captures patterns beyond existing knowledge, or from shortcut learning, where the model exploits spurious correlations. In the former scenario, the model could offer valuable new insights, whereas in the latter scenario, retraining or corrective measures may be necessary.

A limitation of this study, as discussed in Section \ref{Section3.2.4}, is that the ADAGE framework estimates channel group contributions using Channel-Group SHAP approximation. This approximation relies on assumptions of local linearity and independence among channel groups. As a result, the estimated contribution values are inherently approximate, and their direct use as exact measurements should be approached with caution. Another limitation is that the alignment score, while enabling relative comparisons across models, currently lacks a solid theoretical basis as an absolute metric. In addition, uncertainties in the input data, including atmospheric noise, cloud contamination, and sensor-specific artifacts, may propagate into the estimated channel group contributions and potentially lead to misleading interpretations of model behavior. Therefore, the explanation results should be interpreted with appropriate caution, especially when the input data contain substantial uncertainties. Finally, the evaluation did not systematically cover representative climate categories. Therefore, the generalizability of the observed explanation patterns under diverse climatic conditions remains to be validated in future work.

\section{Conclusion}
\label{Section7}

This study presents the ADAGE (Alignment between Domain Knowledge And GeoAI Explanation Evaluation) framework, which systematically evaluates GeoAI model explanations by leveraging domain knowledge from remote sensing, particularly the physical characteristics of spectral bands, in satellite-based flood mapping. The framework employs Channel-Group SHAP approach to estimate the contributions of grouped input channels to pixel-level predictions. By aligning the levels of explanation through channel grouping, this approach enables comparison between GeoAI explanations and domain knowledge. Two satellite-based flood mapping case studies demonstrate the effectiveness of the proposed framework in quantitatively assessing the alignment between pixel-level GeoAI explanations and domain knowledge. In addition, the framework supports the identification of misaligned explanations, which may stem from novel patterns or spurious model behavior. This study contributes to bridging the gap between explainability and domain knowledge in GeoAI for Earth observation, thereby facilitating the adoption of GeoAI models in both scientific research and operational applications. Future work could extend the ADAGE framework to additional Earth observation tasks, such as wildfire detection and landslide mapping, to further assess its generalizability. This may include integrating more diverse forms of domain knowledge, such as temporal dependencies and established physical models. Moreover, developing approaches to refine DLSS-RS models based on misalignments between model explanations and reference explanations could be a promising direction for future research.

\section*{Acknowledgement}

This research was funded in part by the National Science Foundation under award 2120943 and NASA (National Aeronautics and Space Administration) under award 80NSSC24PC446.

\bibliographystyle{cas-model2-names}

\bibliography{ADAGE}

\end{document}